\documentclass{article}

\usepackage[preprint]{neurips_2025}

\usepackage[utf8]{inputenc} 
\usepackage[T1]{fontenc}    
\usepackage{url}            
\usepackage{booktabs}       
\usepackage{amsfonts}       
\usepackage{nicefrac}       
\usepackage{microtype}      
\usepackage{xcolor}         
\usepackage{graphicx}
\usepackage{amsmath}
\usepackage{pifont}
\usepackage{multirow}
\usepackage{makecell}
\usepackage[table]{xcolor}
\usepackage{geometry}
\usepackage{subcaption}
\usepackage{pifont}  
\usepackage{wasysym} 
\usepackage{tikz}
\usepackage{soul}
\usepackage{enumitem}
\usepackage{multirow}
\usepackage[table]{xcolor}
\usepackage{array}
\newcolumntype{L}[1]{>{\columncolor{white}}p{#1}}
\usepackage[pagebackref,breaklinks,colorlinks,citecolor=teal]{hyperref}

\title{\textsc{Care-PD}: A Multi-Site Anonymized Clinical Dataset for Parkinson’s Disease Gait Assessment}

\author{ \small
\renewcommand{\arraystretch}{0.9} 
  \begin{tabular}{@{\hskip 4pt}c@{\hskip 4pt}c@{\hskip 4pt}c@{\hskip 4pt}c}
    Vida Adeli\textsuperscript{1,2,3}\footnotemark[2] 
    &
    Ivan Klabučar\textsuperscript{3} 
    &
    Javad Rajabi\textsuperscript{1,2}
    & 
    Benjamin Filtjens\textsuperscript{2,3} 
    \\
    Soroush Mehraban\textsuperscript{1,2,3}
    &
    Diwei Wang\textsuperscript{4}
    &
    Hyewon Seo\textsuperscript{4}
    &
    Trung-Hieu Hoang\textsuperscript{6} 
    \\
    Minh N. Do\textsuperscript{6}
    &
    Candice Muller\textsuperscript{5} 
    &
    Claudia Neves de Oliveira\textsuperscript{7}
    &
    Daniel Boari Coelho\textsuperscript{7}
    \\
    Pieter Ginis\textsuperscript{8}
    &
    Moran Gilat\textsuperscript{8}
    &
    Alice Nieuwboer\textsuperscript{8} 
    &
    Joke Spildooren\textsuperscript{9} 
    \\
    J. Lucas Mckay\textsuperscript{10}
    &
    Hyeokhyen Kwon\textsuperscript{10} 
    &
    Gari Clifford\textsuperscript{10}
    &
    Christine Esper\textsuperscript{10}
    \\
    Stewart Factor\textsuperscript{10}
    &
    Imari Genias\textsuperscript{10} 
    &
    Amirhossein Dadashzadeh\textsuperscript{11}
    &
    Leia Shum\textsuperscript{3}
    \\
    Alan Whone\textsuperscript{11}
    &
    Majid Mirmehdi\textsuperscript{11}
    &
    Andrea Iaboni\textsuperscript{1,3} 
    &
    Babak Taati\textsuperscript{1,2,3 $\dagger$} 
  \end{tabular} \vspace{8pt}
  \\
\textsuperscript{1}\textit{University of Toronto} \  
\textsuperscript{2}\textit{Vector Institute} \ 
\textsuperscript{3}\textit{KITE Research Institute-UHN} \\  
\textsuperscript{4}\textit{University of Strasbourg}  \
\textsuperscript{5}\textit{University Hospitals of Strasbourg} \\
\textsuperscript{6}\textit{University of Illinois Urbana-Champaign} \
\textsuperscript{7}\textit{Federal University of ABC}\\  
\textsuperscript{8}\textit{KU Leuven}  \
\textsuperscript{9}\textit{Hasselt University} \
\textsuperscript{10}\textit{Emory University} \ 
\textsuperscript{11}\textit{University of Bristol} \ 
}

\begin{document}

\maketitle
\begingroup
\renewcommand\thefootnote{\fnsymbol{footnote}} 
\footnotetext[2]{Corresponding authors: \texttt{\{vadeli, taati\}@cs.toronto.edu}}
\endgroup

\begin{abstract}
Objective gait assessment in Parkinson’s Disease (PD) is limited by the absence of large, diverse, and clinically annotated motion datasets. We introduce \textsc{Care-PD}, the largest publicly available archive of 3D mesh gait data for PD, and the first multi-site collection spanning 9 cohorts from 8 clinical centers. All recordings (RGB video or motion capture) are converted into anonymized SMPL meshes via a harmonized preprocessing pipeline.
\textsc{Care-PD} supports two key benchmarks: supervised clinical score prediction (estimating Unified Parkinson’s Disease Rating Scale, UPDRS, gait scores) and unsupervised motion pretext tasks (2D-to-3D keypoint lifting and full-body 3D reconstruction). Clinical prediction is evaluated under four generalization protocols: within-dataset, cross-dataset, leave-one-dataset-out, and multi-dataset in-domain adaptation.
To assess clinical relevance, we compare state-of-the-art motion encoders with a traditional gait-feature baseline, finding that encoders consistently outperform handcrafted features. Pretraining on \textsc{Care-PD} reduces MPJPE (from 60.8\,mm to 7.5\,mm) and boosts PD severity macro-F1 by 17 percentage points, underscoring the value of clinically curated, diverse training data. \textsc{Care-PD} and all benchmark code are released for non-commercial research at \href{https://neurips2025.care-pd.ca/}{\texttt{\small https://neurips2025.care-pd.ca}}. 
\end{abstract}

\section{Introduction}
Accurate gait assessment is essential for PD diagnosis, monitoring, and treatment planning; but current clinical evaluations remain subjective and hard to scale. While automated motion analysis offers objective, reproducible metrics, progress is hindered by small, single-site datasets lacking standardization. There is a critical need for large, diverse, and \emph{publicly available} motion datasets with clinical labels to enable generalizable machine learning models for real-world use.

We introduce \textsc{Care-PD}, a multi-institutional dataset aggregating nine datasets from eight clinical sites, encompassing optical motion-capture and RGB video data. All sequences are converted to a unified 3D volumetric representation~\cite{loper2023smpl} using a reproducible preprocessing pipeline that includes data cleaning, temporal segmentation, sensor harmonization, and privacy-preserving SMPL mesh conversion. The anonymization process ensures compliance with institutional ethics and suitability for public research use.
\textsc{Care-PD} is the first clinically annotated motion dataset of this scale focused on parkinsonian gait. Over one-third of the walks include clinician-rated UPDRS-gait scores, referring to the gait-specific item of the Unified Parkinson’s Disease Rating Scale~\cite{goetz2008movement} motor examination. The remaining data include other clinical attributes---e.g., medication status, Freezing-of-Gait (FoG) presence, PD diagnosis---or are unlabeled for pretraining and self-supervised learning.

Alongside the dataset, we release benchmark protocols for two tasks: (1)~clinical severity estimation from gait sequences, and (2)~motion pretext tasks, including 2D-to-3D lifting and 3D reconstruction. These tasks evaluate both the clinical utility of pretrained motion encoders and the benefit of incorporating clinically grounded data into self-supervised training. Our experiments show that while state-of-the-art encoders pretrained on generic motion retain \emph{some} latent structure useful for clinical prediction, exposure to pathological gait in \textsc{Care-PD} substantially improves their reconstruction quality and downstream severity estimation.

\textsc{Care-PD} is publicly released under a research-only license, following privacy-preserving protocols and institutional approvals. It serves as a valuable resource for developing machine learning methods that bridge general-purpose motion modeling and clinical utility in PD care, and offers a testbed for evaluating representation learning, domain adaptation, and clinical motion understanding tasks.

\vspace{-7pt}\section{Related Work}
\vspace{-7pt}\paragraph{Gait Assessment.} 
Recent machine learning approaches for Parkinsonism gait assessment span objectives such as diagnostic classification~\cite{juutinen2020parkinson, rehman2019selecting, ma2024twin, huo2025early}, clinical severity scoring~\cite{han2023automatic, eguchi2023gait, navita2025gait, tian2024cross, adeli2024benchmarking, endo2022gaitforemer, sabo2020assessment, sabo2022estimating}, FoG detection~\cite{sigcha2020deep, abbasi2025deep, bikias2021deepfog, filtjens2022automated}, and symptom measurement like bradykinesia~\cite{sama2017estimating} or tremor~\cite{duque2024deep}. While methods using wearables or video show high accuracy and strong correlations with clinical ratings~\cite{heye2024validation}, they are typically limited by small, single-site datasets and perform poorly under real-world variability~\cite{di2020gait, sabo2023evaluating}. Similarly, FoG models~\cite{sigcha2020deep, abbasi2025deep, bikias2021deepfog, filtjens2022automated} and motor symptom assessments~\cite{deng2024interpretable} are largely validated in lab settings with task-specific protocols, limiting their generalizability~\cite{mancini2021measuring, mancini2025technology}. These challenges underscore the need for large, diverse datasets to support robust and generalizable Parkinsonism assessment.

\vspace{-7pt}\paragraph{General Motion Benchmarks.} 
Large-scale datasets like Human3.6M~\cite{ionescu2013human3}, AMASS~\cite{mahmood2019amass}, and NTU RGB+D~\cite{shahroudy2016ntu} offer extensive motion, RGB, and depth data from healthy individuals performing general activities. Although they include walking, they are not designed for clinical tasks and lack pathological gait patterns, limiting their usefulness for applications such as disease severity estimation or gait abnormality detection.

\vspace{-7pt}\paragraph{Gait Datasets.} 
Parkinsonism gait has been studied using a wide range of modalities, including inertial sensors~\cite{DaphnetFreezingOfGait, bot2016mpower, anderson2024weargait, morgan2023multimodal, ribeiro2022public}, pressure platforms~\cite{chatzaki2021smart, anderson2024weargait, hausdorff2000dynamic}, optical motion capture~\cite{shida2023public, warmerdam2022full, morgan2023multimodal}, RGB video~\cite{kim2024tulip, kour2020gait, he2022integrated, ribeiro2022public, deng2024interpretable}, multimodal physiological sensors~\cite{zhang2022multimodal, abtahi2020merging}, and radio-frequency (RF) methods~\cite{zhang2024mp}. Each modality has distinct strengths and trade-offs: IMUs offer portability but lack anatomical context~\cite{guo2022detection}; pressure platforms provide spatiotemporal data but are lab-bound and anatomically limited; optical systems offer high biomechanical fidelity but lack scalability~\cite{guo2022detection, gurbuz2024overview}; video suffers from occlusions and limited accuracy; physiological sensors require complex setups~\cite{zhang2022multimodal}; and RF methods, while unobtrusive, lack anatomical resolution.
To address the scalability challenges of prior work, this study merges RGB video and optical motion capture using the SMPL-based representation, which unifies their differing content and marker conventions within a common anatomical framework.
\section{\textsc{Care-PD}  Dataset}
\label{sec:dataset}
\textsc{Care-PD} is a clinically grounded dataset for automated PD gait assessment, unifying heterogeneous gait recordings from multiple clinical sites into a standardized 3D mesh format using SMPL. As the largest publicly available collection of 3D body meshes focused on PD gait, it supports model development and evaluation for PD severity estimation and related motion analysis tasks.

\setlength{\tabcolsep}{1.5pt}
\begin{table}
  \caption{\small 
    \textsc{Care-PD}  dataset overview. “RGB” = monocular video; “MoCap” = optical-marker motion capture. “FPS” is the original frame or marker rate. “Duration” is the total time (in minutes) retained after gait-segment extraction. “Med” = medication state (on/off), “PD/HC” = Parkinson’s vs. healthy control, “FoG” = subject-level freezing of gait (freezing/Non-freezing) label. “$\pm$” indicates (mean $\pm$ std). 
    }
  \label{cohort}
  \centering 
  {\fontsize{9pt}{10pt}\selectfont
  \begin{tabular}{lcccccccc}
    \toprule
    Sub-dataset     &  \makecell{Orig. \\ Modality} & \makecell{Orig. \\ FPS} & \#Subjects & \#Walks & \makecell{Duration \\ (min:sec)} &  \makecell{Age \\ (years)} & \makecell{Sex \\ (\%male)} & Annotation\\
    \midrule
    PD-GaM~\cite{dadashzadeh2024pecop}  & RGB & 25 & 30 & 1701 & 186:22 & 54.1 $\pm$ 8.1& 56.7 & UPDRS-gait \\
    BMClab~\cite{shida2023public}  & MoCap & 150 & 23 & 781 & 46:57 & 65.6 $\pm$ 8.3 & 78.3 & UPDRS-gait + FoG + Med \\
    T-SDU-PD & RGB & 30 & 14 & 381 & 49:39 & 76.2 $\pm$ 8.7 & 57.1 & UPDRS-gait \\
    3DGait~\cite{wang2023video}  & RGB & 30 & 43 & 90 & 14:59 & 78.43 $\pm$ 9.3 & 30.2 & UPDRS-gait \\
    KUL-DT-T~\cite{spildooren2010freezing}  & MoCap & 100 & 29 & 763 & 64:45 &  65.2 $\pm$ 6.8 & 79.3 & FoG  \\
    DNE~\cite{hoang2024smartphone}  & RGB & 60 & 97 & 476 & 21:27 &64.1 $\pm$ 14.3 & 48.7& PD/HC \\
    E-LC~\cite{lucas2019freezing}  & MoCap & 120 & 59 & 162 & 202:32 & 68.2 $\pm$ 8.3 & 76.3 & FoG + Med \\
    T-SDU & RGB & 30 & 53 & 2799 & 341:09 & 77.1 $\pm$ 8.0&  57.4 & -  \\
    T-LTC & RGB & 30 & 14 & 1324 & 196:04 & 81.4 $\pm$ 6.7 & 28.6 & -  \\
    \midrule
    Total & - & - & 362 & 8477 & 1123:54 & 70.0 $\pm$ 8.7  & 56.9 & - \\

    \bottomrule
  \end{tabular} \vspace{-15pt}
  }
\end{table} 

\subsection{Participating Sites and Cohorts} 
\label{sec:sites}

\textsc{Care-PD} aggregates gait recordings from 9 studies across 8 clinical centres in 6 countries, all collected under local  institutional review board approval with written informed consent. Retrospective analysis of these existing datasets was approved by the Social Sciences, Humanities \& Education Research Ethics Board of the University of Toronto (REB \#47891). The cohorts differ in population, environment, and capture setup, offering diversity for training and evaluating generalizable models.
Table~\ref{cohort} summarizes key metadata, including modality, frame rate, subject count, duration, and clinical annotations. Four datasets include UPDRS-gait scores (0–3), though score 3—reflecting severe impairment and often requiring assistive devices—is rare and appears only in PD-GaM and 3DGait. Further score distribution details are provided in the Appendix~\ref{appendix:datasetdetail}.

The T-SDU and T-LTC datasets (\textit{RGB}) were collected in prospective observational studies on gait changes and fall risk~\cite{sabo2020assessment, mehdizadeh2020vision} from inpatient participants at a specialized dementia unit and a long-term care facility in Toronto, Canada. These datasets are being publicly released for the first time in this work, following written informed consent and ethical approval from the University Health Network Research Ethics Board (CAPCR ID 24-5835). A 14-subject subset, T-SDU-PD, was selected to capture diverse parkinsonian gait patterns and annotated with  per walk UPDRS-gait by expert clinicians. Gait data were recorded via a ceiling-mounted camera triggered by an RFID every time participants walked naturally through a hallway. For inclusion in \textsc{Care-PD}, recordings were curated to retain only clean walking segments, excluding turns, stops, and other non-walking behaviors.

PD-GaM  (\textit{RGB})~\cite{adeli2025gaitgen} is a 3D mesh gait dataset derived from PD4T (University of Bristol, UK)~\cite{dadashzadeh2024pecop}, comprising per walk UPDRS-scored trials from four PD motor-function tasks. \textsc{Care-PD}  includes the gait task subset, where each trial captures a participant walking back and forth resulting in four walking segments per recording.

The DNE dataset (\textit{RGB})~\cite{hoang2024smartphone, 9760105} was collected across multiple clinical sites in the United States, including OSF HealthCare (Illinois), Bradley Physical Therapy (Washington, Pennsylvania), and Bon Secours St. Francis Inpatient Rehabilitation Center (South Carolina), as part of a neurological assessment study using smartphone video recordings. It includes recordings of participants performing up to five standardized tasks, including fine-motor, facial, and gait assessments. For \textsc{Care-PD}, we include the stand-up and walk task subset, where participants walked back and forth with labels assigned per walk. Clean walking segments were extracted for inclusion, similar to PD-GaM.

The 3DGait dataset (\textit{RGB})~\cite{wang2023video} was collected at the University of Strasbourg (France) to support gait analysis in neurodegenerative diseases. It includes gait videos from individuals with Alzheimer's disease, dementia with Lewy bodies, and  healthy controls, all assessed using the per walk MDS-UPDRS-gait criterion. Recordings were made during clinical exams using an RGB camera as patients walked across an 8-meter GAITRite walkway, with alternating front, back, or side views. \textsc{Care-PD} includes the subset of trials featuring UPDRS-labeled straight walks.

The BMClab dataset (\textit{MoCap})~\cite{shida2023public} was collected in the Laboratory of Biomechanics and Motor Control, Federal University of ABC~(Brazil), using a Raptor-4 optical motion-capture system (Motion Analysis Corp.) with 44 reflective markers placed following a clinical full-body protocol. It includes walking trials from PD participants with and without FoG, in both ON- and OFF-medication, with MDS-UPDRS-gait labels per participant in each medication state by expert clinicians.

The KUL-DT-T dataset (\textit{MoCap})~\cite{spildooren2010freezing,filtjens2022automated} was recorded in the Movement Disorders Clinic of the University Hospital Leuven~(Belgium), using a 8-camera 3D optical motion capture system with 34 markers (Vicon Motion Systems), under dual-task and turning conditions designed to provoke FoG. It includes participants with and without FoG.

The E-LC dataset (\textit{MoCap})~\cite{lucas2019freezing, kwon2023explainable} was acquired at the Emory Movement Disorders Clinic~(Atlanta, USA) using a 14-camera 3D optical motion capture system (Motion Analysis Corp.) with 60 reflective markers, capturing high-resolution data from PD participants across medication states (ON/OFF) per walk and FoG subtypes per participant under standardized protocols.

\vspace{-5pt}\subsection{Data Harmonization}
\label{sec:harmonization} \vspace{-5pt}

\textsc{Care-PD} harmonizes heterogeneous gait data using a preprocessing pipeline that converts all recordings into 30~Hz 3D SMPL mesh sequences, incorporating modality-specific processing, artifact correction, anonymization, and subject-level stratified evaluation splits.

\vspace{-7pt}\paragraph{MoCap Processing.} 
MoCap data underwent: 1)~Quality control to fix joint errors from dropped or noisy markers; 2)~Joint standardization by mapping each system’s layout to 22 SMPL joints; 3)~SMPL fitting using SparseFusion optimization~\cite{zuo2020sparsefusion} to estimate full-body parameters from sparse 3D joints; and 4)~Downsampling to 30~Hz and removal of corrupted or very short segments to ensure consistent, anatomically valid outputs.

\vspace{-10pt}\paragraph{Video Processing.}
For RGB datasets: 1)~SMPL meshes were extracted using WHAM~\cite{wham:cvpr:2024}, a monocular mesh recovery method (see Appendix~\ref{appendix:datasetdetail} for clinical validity experiments); 2)~The extracted mesh was visually verified to correspond to the intended patient
when multiple people appeared in a frame; 3)~Only clean walking segments were retained by removing non-walking behaviors; and 4) A slope correction using the Kabsch algorithm~\cite{lawrence2019purely} was applied to counteract distortions from ceiling-mounted cameras, aligning recovered 3D walks to a canonical ground plane and preserving gait dynamics while correcting the camera-induced skew (see Appendix~\ref{appendix:datasetdetail}).
\vspace{-10pt}\paragraph{Anonymization.} 
To ensure privacy, only textureless SMPL mesh parameters are released—excluding video frames, identifiable 3D point clouds, and potentially identity-revealing SMPL shape parameters. For longitudinal datasets (T-SDU, L-SDU, and T-SDU-PD), start dates are standardized to anonymize timelines, and all subject IDs and filenames are anonymized.

The resulting dataset contains 18.66 hours of anonymized 3D gait-mesh sequences, totaling 8,477 walking segments.
We resample every recording to a common \(\sim\)30~FPS and provide subject-level train/val/test splits stratified by key UPDRS-gait scores, with multiple configurations depending on the evaluation protocol. Full label distributions, per-split statistics, and additional preprocessing details are included in Appendix~\ref{appendix:datasetdetail}.

\begin{figure}
    \centering
    \includegraphics[width=\linewidth]{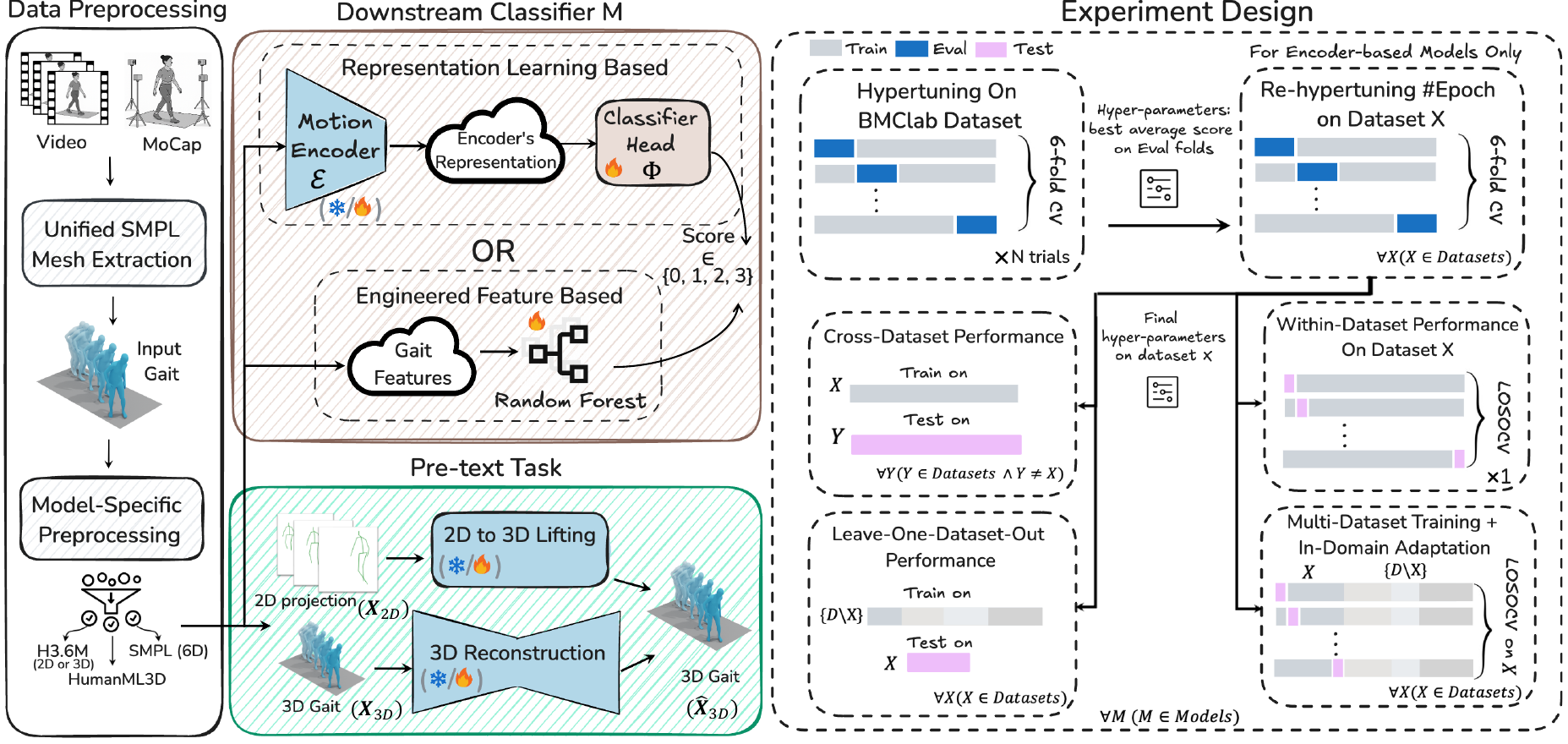}
    \caption{\small Overview of \textsc{Care-PD} preprocessing and experimental design. Left: Unified pipeline for extracting SMPL gait meshes from MoCap and video data, followed by model-specific formatting. Right: Benchmarking setup across two pipelines (representation learning vs. gait features), pretext tasks, and four evaluation protocols.}
    \label{fig:diagram} \vspace{-15pt}
\end{figure}
\vspace{-7pt}\section{Benchmarks \& Experiments}

\subsection{Clinical Score Estimation Task}
A walk is represented as a sequence of body pose frames {\small$\mathbf{M}^{1:T} = \{ \mathbf{m}^1, \mathbf{m}^2, \ldots, \mathbf{m}^T \}$},
where each frame {\small\( \mathbf{m}^t \in \mathbb{R}^F \)} encodes the pose at time \( t \) using \( F \) parameters. 
Given a walk {\small\( \mathbf{M}^{1:T} \)}, the objective is to estimate its associated UPDRS-gait severity score \( S \in \{0,1,2,3\} \) reflecting the degree of gait impairment.
We evaluate two baseline approaches for this task: \textit{Representation-learning}, using deep encoders trained on motion sequences, and \textit{Engineered gait features}, using traditional handcrafted features with classical classifiers. These baselines help assess the trade-offs between learned and interpretable features, and analyze their sensitivity to clinical labels across different data sources.

\vspace{-7pt}\paragraph{Representation-Learning Baselines.}
To predict a person’s UPDRS-gait score \( S \) from their walk {\small\( \mathbf{M}^{1:T} \)} using a motion encoder \( \mathcal{E} \), we first apply a series of preprocessing steps \( P_{\mathcal{E}} \) to convert variable-length input into several non-overlapping motion clips of fixed length \( N \), i.e., {\small$\{ \mathbf{p}^{1:N}_1, \ldots, \mathbf{p}^{1:N}_{\frac{T}{N}} \} = P_{\mathcal{E}}(\mathbf{M}^{1:T})$}.
The preprocessing steps are encoder-specific and include operations such as windowing, normalization, and format conversion, e.g., from SMPL to different joint coordinate format. Full details of these steps for each encoder are provided in Appendix~\ref{appendix:datasetdetail}.
Each motion clip \( \mathbf{p}_i \) is fed into the encoder \( \mathcal{E} \) to obtain a latent feature vector \( \mathbf{e}_i \), which is then passed to a lightweight classifier head \( \Phi \) to produce a predicted score {\small$ s_i = \Phi(\mathcal{E}(\mathbf{p}_i)) $}.
The final predicted score \( S \) for the entire sequence {\small\( \mathbf{M}^{1:T} \)} is obtained via majority voting across all predictions \( \{ s_i \} \). This approach enables score prediction even on long recordings by aggregating local clip-level predictions.

We deliberately keep each motion encoder trained on its original task frozen and evaluate it using two lightweight probes (a linear classifier and a k-nearest neighbors (k-NN) classifier) to examine whether state-of-the-art motion encoders, pre-trained on generic human motion data, already capture clinically relevant gait features.
This setup allows us to investigate three key aspects:
1) Clinical usefulness: We compare pretrained encoders with traditional gait features to determine if their representations, when paired with simple probes, can match or outperform traditional approaches, quantifying their clinical utility.
2) Generalizability and out-of-distribution robustness: Subjects in Care-PD exhibit characteristics not typically seen in the encoders' pretraining on healthy human motion datasets (e.g., parkinsonian shuffling, tremors, or subtle asymmetries). We investigate whether these clinical patterns are effectively represented in the latent space or instead disregarded as noise by the pretrained encoders.
3) Real-world deployability: Many clinical settings cannot support full model fine-tuning. Frozen probes simulate “plug-and-play” usage, where public models are applied directly to clinical tasks with minimal adaptation.

We evaluate seven state-of-the-art validated motion encoders: POTR~\cite{martinez2021pose}, MixSTE~\cite{zhang2022mixste}, PoseFormerV2~\cite{zhao2023poseformerv2}, MotionBERT~\cite{zhu2023motionbert}, MotionAGformer~\cite{mehraban2024motionagformer},  MotionCLIP~\cite{tevet2022motionclip}, and MoMask~\cite{guo2024momask}, spanning a diverse range of human motion tasks including 2D-to-3D lifting, 3D pose reconstruction, action recognition, and motion generation. 
Further details on the encoders, encoder-specific preprocessings and motion representations used are provided in Appendix~\ref{appendix:datasetdetail}.

\vspace{-10pt}\paragraph{Engineered-Feature Baseline.}
As a classical baseline, we train a Random Forest classifier on a set of interpretable gait features derived from the reconstructed body joints. These features include spatiotemporal descriptors (e.g., cadence, step length, step width, step time, walking speed)~\cite{zanardi2021gait, sabo2020assessment}, stability-related measures (e.g., estimated margin of stability)~\cite{watson2021use}, and motor/postural indicators (e.g., foot lifting, arm swing, stoop posture)~\cite{mirelman2016arm, yoon2019effects}. Many of these directly correspond to criteria used in the UPDRS-gait scoring rubric~\cite{goetz2008movement}.
Random Forest models built on such handcrafted variables have shown to outperform several other machine-learning algorithms in differentiating PD from control gait~\cite{munoz2022machine}. We also tested kernel SVM and XGBoost, but Random Forest consistently yielded comparable or better results. For simplicity, we report only the Random Forest baseline. Full details on feature extraction are provided in Appendix~\ref{gaitfeat-app}.

\vspace{-7pt}\subsection{Evaluation Protocols}
To understand how well different representations support clinical score estimation, we design four evaluation protocols that reflect realistic deployment scenarios (See Fig.~\ref{fig:diagram}).

\vspace{-10pt}\paragraph{Within and Cross Dataset Evaluation.}
Each dataset in \textsc{Care-PD} is unique in terms of recording setup, capture geometry, and population demographics and movement instructions. To evaluate model performance under these conditions, we assess both within-dataset and cross-dataset generalization.
\textit{Within-dataset evaluation} uses a Leave-One-Subject-Out (LOSO) cross-validation strategy to ensure no subject-specific motion patterns leak between splits.
\textit{Cross-dataset evaluation} tests how well models trained on \emph{one} dataset transfer to others.  We train a classifier on a single dataset and evaluate it on the remaining ones.
This simulates deployment in a new clinical environment without access to site-specific labeled data and highlights the model’s robustness to changes in protocol, hardware, and patient characteristics.
Together, these two protocols reveal the extent to which learned representations are sensitive to dataset-specific biases and whether they generalize across both subjects and sites.

\vspace{-10pt}\paragraph{Leave One Dataset Out (LODO).}
Clinical sets are small and often carry cohort‐specific biases, which can cause classifiers to overfit to a single dataset.  
To test whether diversity can dilute these biases, we train on the union of {\small\(D-1\)} cohorts and evaluate on the held-out cohort.  
A gain over the single‐cohort cross‐test of the previous section indicates that combining heterogeneous data lets the probe focus on pathology‐related variation rather than spurious site cues.  
Conversely, a persistent gap reveals that the held-out cohort contains systematic differences that even large, diverse training data cannot bridge, highlighting where additional harmonization or domain adaptation is warranted.

\vspace{-10pt}\paragraph{Multi-dataset In-domain Adaptation (MIDA).}
While LODO tests generalization to unseen domains, many real-world deployments allow limited access to target-domain data. MIDA explores whether using a small amount of in-domain data can significantly boost performance.
Starting from the LODO checkpoint, we fine-tune the probe (but keep the encoder frozen) on the target cohort’s training split—again under LOSO—and test on its held-out subjects.  
Comparing MIDA to LODO quantifies how much performance can be recovered by a modest amount of in-domain supervision.

\vspace{-10pt}\subsection{Training \& Metrics}
We tuned classifier head hyperparameters using 6-fold stratified cross-validation on the BMClab dataset, selected for its size and clean motion quality. The best combination was applied across all datasets, adjusting only the number of training epochs per dataset.
Full tuning details, search space, and evaluation strategy are in Appendix~\ref{app:hyperparams}.
UPDRS-gait score 3 (indicating severe impairment) is rare and is present in only two datasets (PD-GaM and 3DGait). In cross-dataset evaluations, it may be entirely absent from the training set but present in the test set. In such cases, the classifier is unable to predict label 3, resulting in an F1 score of 0 for that class. Therefore, including it in macro averaging artificially deflates the overall metric and make comparisons unfair.
To address this, we report macro F1 scores in two setups: including all UPDRS-gait labels {0, 1, 2, 3} ($\text{F}1_{0-3}$), and excluding label 3 ($\text{F}1_{0-2}$). 
Importantly, all score-3 samples remain in the training folds; the exclusion applies only to the metric.
We adopt macro (unweighted) averaging so that each class contributes equally despite imbalanced class distributions, and we choose F1 to balance precision and recall without favoring either. We report results using the linear probe, as it showed no substantial difference from the k-NN.

\vspace{-10pt}\subsection{Motion Pre-text Tasks}
To evaluate the utility of \textsc{Care-PD} in improving motion representation learning and 3D pose estimation, we conducted experiments on two common pretext tasks: 2D-to-3D lifting and 3D reconstruction.
We used two top-performing models: MotionAGFormer~\cite{mehraban2024motionagformer} for 2D lifting and MoMask~\cite{guo2024momask} for motion reconstruction and generation. Both models were pre-trained on generic, able-bodied motion datasets. 
We then fine-tuned them or trained them from scratch on \textsc{Care-PD} to assess: (1) Clinical adaptation impact on proxy accuracy: whether exposure to \textasciitilde19 hours of diverse pathological gait improves 3D estimation error measured via mean per joint position error (MPJPE), Procrustes-aligned MPJPE (PA-MPJPE), and acceleration error (Acc) common for the task~\cite{kanazawa2019learning}. (2) Downstream impact: whether such improvements translate to better downstream UPDRS-score prediction, when coupled with the same lightweight probe used in the clinical task.
This experiment tells us whether injecting clinically rich motion into SOTA encoders (i) improves 3D pose accuracy on pathological gait, and (ii) yields clinical-task benefits \emph{without} extra model capacity.

\vspace{-5pt}\section{Results \& Analysis}
\begin{figure}[t]
    \centering
    \includegraphics[width=.8\linewidth]{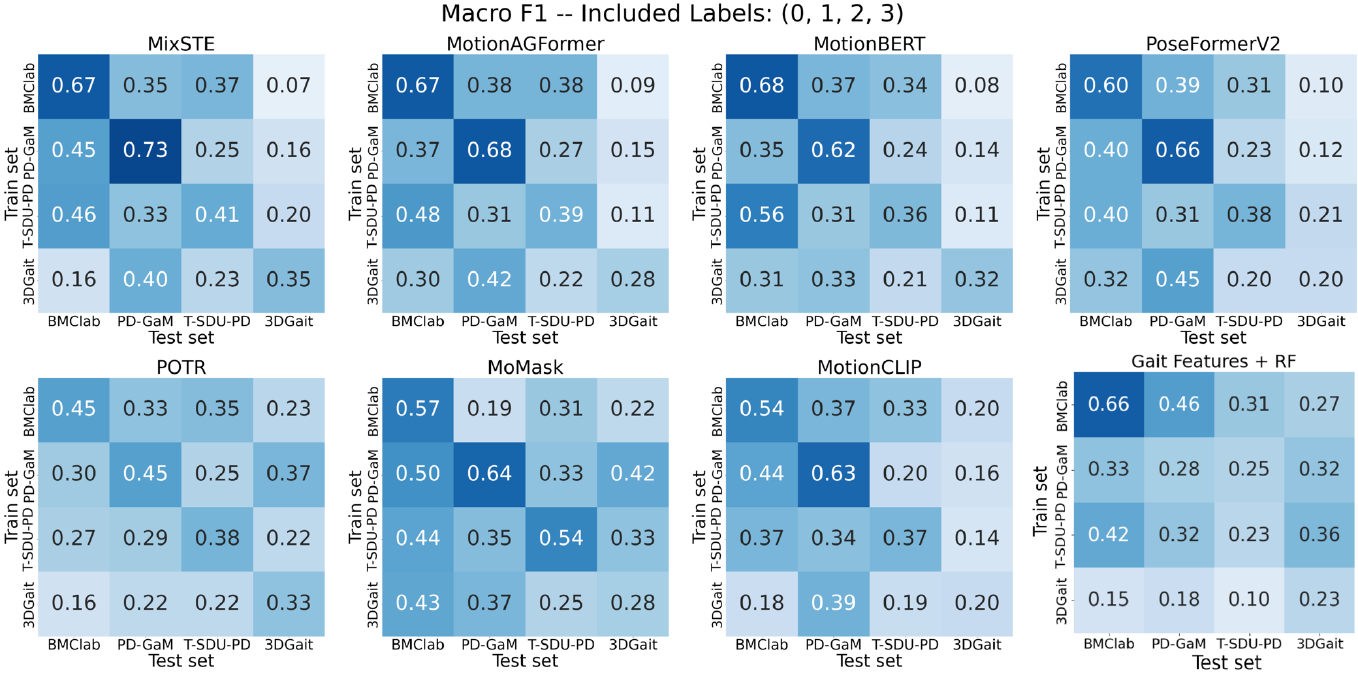} \vspace{-5pt}
    \caption{\small Within-dataset and cross-dataset macro-$\text{F}1_{0-3}$ scores for encoder and gait features-based models.}
    \label{fig:combined_within_cross_4class} \vspace{-10pt}
\end{figure}

\begin{figure}[t]
    \centering
    \includegraphics[width=.8\linewidth]{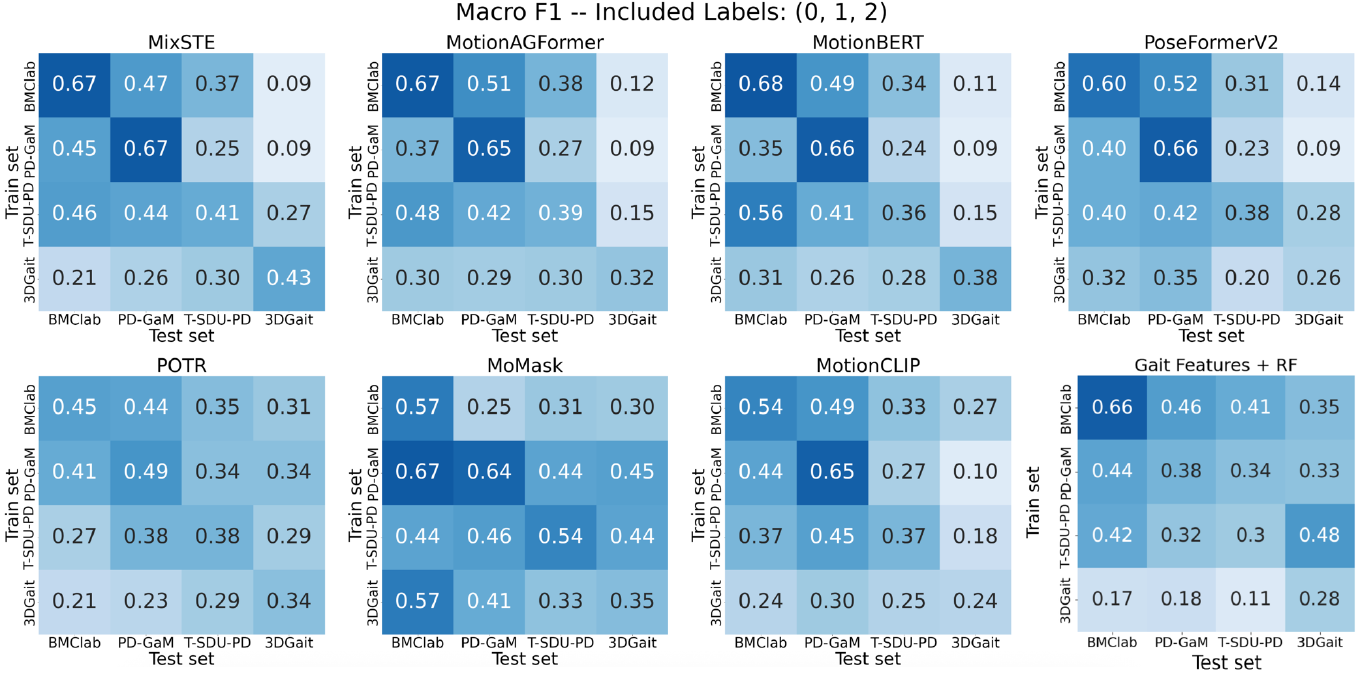}
    \caption{\small Cross-dataset and within-dataset macro-$\text{F}1_{0-2}$ scores using encoder-based models.}
    \label{fig:combined_within_cross_3class} \vspace{-15pt}
\end{figure}

\subsection{Severity Estimation Benchmarks}
\noindent\textbf{Within and Cross Dataset.} We evaluate model performance using within-dataset (LOSOCV; diagonal) and cross-dataset (off-diagonal) protocols, reporting $\text{F}1_{0-3}$ and $\text{F}1_{0-2}$ in Fig.~\ref{fig:combined_within_cross_4class} and Fig.~\ref{fig:combined_within_cross_3class}, respectively. 
Most encoders achieve strong in-site performance on large cohorts—up to 0.73 on PD-GaM and 0.68 on BMClab—demonstrating that frozen representations retain some clinically relevant information. Performance drops for the smaller T-SDU-PD set and falls further for 3DGait. 
Transferring to unseen datasets typically reduces F1 by 0.2 to 0.4, exposing domain gaps driven by data distribution shifts. Models trained on PD-GaM generalize best, likely due to its scale and diversity. Among all the backbones, the VQ-VAE MoMask prove the most robust: when it is trained on PD-GaM, its average cross-site F1 remains above 0.40, in several cases matching or surpassing the within-site result of weaker encoders (see Appendix ~\ref{app:app-exp}). Removing class 3 from evaluation consistently boosts metrics, especially on small datasets, confirming its sparsity and ambiguity.

Compared to these, the handcrafted-feature baseline underperforms on most cross-dataset settings, highlighting the superior portability of learned motion representations. Overall, the results underscore the value of multi-site evaluation and reveal the need for models that generalize beyond their training domain.
The results validate our multi-site evaluation protocol: single-cohort scores over-estimate readiness for deployment, while cross-site tests reveal both the promise of modern encoders and the persistent domain gaps that future work must bridge.

\begin{figure}[t]
    \centering
    \includegraphics[width=1\linewidth]{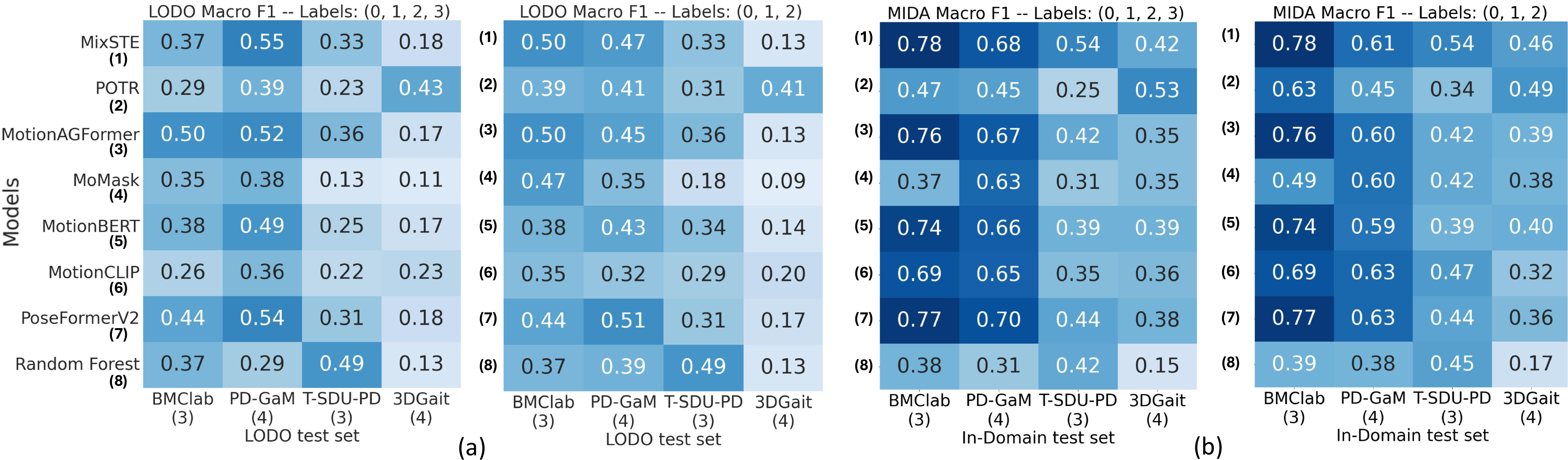}
    \caption{\small Macro-$\text{F}1_{0-3}$ and Macro-$\text{F}1_{0-2}$ scores for LODO (left two blocks) and MIDA (right two blocks) evaluations, comparing severity estimation across models and datasets.}
    \label{fig:f1_lodo_mida}
\end{figure}

\noindent\textbf{LODO Analysis.}
In the LODO protocol (Fig.~\ref{fig:f1_lodo_mida}\textcolor{red}{-a}) we observe a clear degradation in performance compared to within-dataset training, confirming that domain shift remains a major challenge for generalization. The two most diverse target sets (BMClab and PD-GaM) are now handled best: MixSTE and MotionAGFormer reach macro-F1 $\approx$ 0.50 on both label configurations, with PoseFormerV2 close behind. By contrast, the small 3DGait cohort stays difficult for every deep backbone ($\leq$  0.18). Across backbones, MotionBERT is the most sensitive to the choice of target site. It peaks at 0.49 on PD-GaM but slips to 0.25 on T-SDU-PD, whereas MotionAGFormer and PoseFormerV2 yield the most consistent scores. Including the rare class 3 generally lowers every entry by 2–5 percentage points, yet leaves the relative ordering intact. 
The low performance in 3DGait is likely due to its limited size—only 90 videos from 43 participants—causing the LOSO setup to leave just one or two test samples per fold, making evaluation unstable.
The Random Forest baseline remains surprisingly competitive on T-SDU-PD, especially in the 3-class setup (0.49), but its performance deteriorates in PD-GaM and 3DGait, indicating poor robustness outside the training domain.

\noindent\textbf{MIDA analysis.} 
The MIDA protocol yields higher F1 scores across the board (Fig.~\ref{fig:f1_lodo_mida}\textcolor{red}{-b}), confirming that augmenting training with a mix of diverse source domains and further tuning on the target significantly boosts performance and adaptation across nearly all settings. On BMClab, all backbones now exceeds 0.69, with MixSTE, MotionAGFormer, MotionBERT, and PoseFormerV2 reaching 0.74 to 0.78, halving the LODO error. PD-GaM shows similar gains, with the same models achieving 0.63–0.70 despite the inclusion of the more ambiguous class 3. Smaller datasets like T-SDU-PD also benefit: MotionAGFormer, MixSTE, and PoseFormerV2 improve by \textasciitilde0.25 absolute F1 over their LODO scores. 
Comparing MIDA (Fig.~\ref{fig:f1_lodo_mida}\textcolor{red}{-b}) to within-dataset LOSO (diagonal elements of Fig.~\ref{fig:combined_within_cross_4class} and Fig.~\ref{fig:combined_within_cross_3class}) 
highlights the value of the CARE-PD dataset in boosting performance. For instance, MixSTE's macro-F1 on BMCLab improves from 0.67 to 0.78 (see Appendix~\ref{app:app-exp}).

    \setlength{\tabcolsep}{2pt}
    \begin{table}\vspace{-10pt}
      \caption{\small Impact of CARE-PD on motion pretext tasks and downstream severity prediction.}
    \label{tab:pretext}
      \label{Pre-training}
      \centering \scriptsize
      \begin{tabular}{llccccccc}
        \toprule
        Model & Task & Train Data & Finetune Data & MPJPE~$\downarrow$\tiny{(mm)} & PA-MPJPE~$\downarrow$\tiny{(mm)} & Acc~$\downarrow$\tiny{(mm/$\text{s}^2$)} & F1-score~$\uparrow$\\
        \midrule
        \multirow{4}{*}{MotionAGFormer~\cite{mehraban2024motionagformer}} 
            & \multirow{4}{*}{2D-3D lifting} 
            & H3.6M    & -            & 60.7 & 21.4 & 99.8 & 48.1 \\
            &                         & H3.6M    & Healty Gait & 29.8 & 7.3  & 35.4  & 50.1  \\
            &                         & H3.6M    & \textbf{\textsc{Care-PD}}      & \textbf{7.5} & \textbf{2.6} & \textbf{11.6} & \textbf{65.1} \\
            &                         & \textbf{\textsc{Care-PD}}   & -           &    \underline{9.0}   &    \underline{3.2}   &    \underline{13.8}     &   \underline{62.3}    \\
        \midrule
        \multirow{4}{*}{MoMask~\cite{guo2024momask}} 
            & \multirow{4}{*}{3D reconst.} 
            & HumanML3D & -           & 22.5& 17.8 & 4.3 &   41.4    \\
            & & HumanML3D & Healty Gait & 22.3 & 13.7 & 4.4 &   40.6    \\
            & & HumanML3D & \textbf{\textsc{Care-PD}}& \textbf{8.7}& \textbf{6.3}& \textbf{2.2} &  \textbf{62.7}     \\
            & & \textbf{\textsc{Care-PD}}    & -     & \underline{9.6}& \underline{7.3}& \underline{3.3} &  \underline{59.8}     \\
        \bottomrule
      \end{tabular}
    \end{table}

\vspace{-10pt}\subsection{Motion Pre-text Results}
For each of the two pretext tasks, we compare four training regimes:
1) Zero-shot: pre-trained on generic motion datasets (H3.6M for MotionAGFormer, HumanML3D for MoMask) and evaluated directly on \textsc{Care-PD}. 2) Fine-tuning on \textsc{Care-PD}. 3) Healthy-gait fine-tuning: fine-tuned instead on 7,971 healthy gait clips from~\cite{schreiber2019multimodal, santos2022multi, bertaux2022gait, grouvel2023dataset} datasets to isolate the impact of merely focusing on walking. 4) Training from scratch on \textsc{Care-PD} without any external pre-training.

Testing is performed exclusively on \textsc{Care-PD}. For each of the 9 \textsc{Care-PD} datasets, we performed a separate 80/20 subject-stratified split (identical across all four training regimes) and report aggregate test error over all test splits. This pooled setting is considerably more challenging than the dataset-specific benchmarks used in previous sections, making the improvements especially meaningful.

Results in Tab.~\ref{tab:pretext} show that fine-tuning on \textsc{Care-PD} significantly boosts both reconstruction metrics across \emph{all} metrics and downstream clinical score prediction.
MPJPE dropped from 60.7 to 7.5 for MotionAGFormer and from 22.5 to 8.7 for MoMask, while the UPDRS-gait F1-score improved from 48.1 to 65.1 and from 41.4 to 62.7, respectively.

Notably, fine-tuning on healthy walks delivers far smaller gains, confirming that the improvement stems from exposure to pathological kinematics rather than from seeing more walking alone. Training from scratch on \textsc{Care-PD} achieves comparable results, suggesting that the dataset is sufficiently rich to support effective learning of subtle biomechanical distinctions in pathological gait, yet external pre-training on diverse motions still leads to lower 3D reconstruction error.

\begin{figure}
    \centering
    \includegraphics[width=0.7\linewidth]{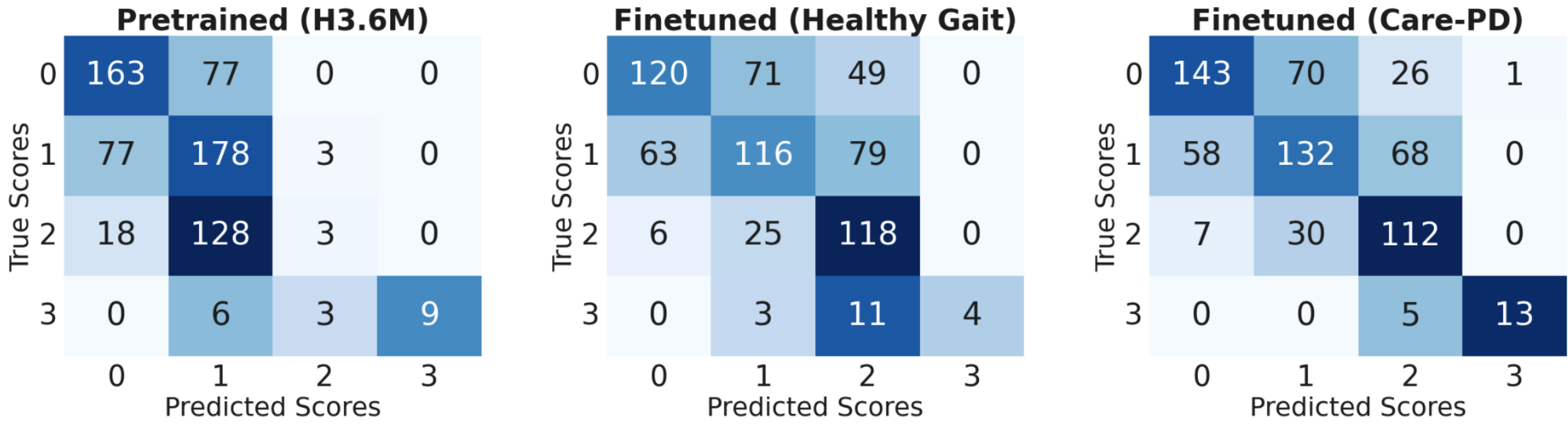}
    \caption{\small MotionAGFormer UPDRS-gait confusion matrices under different pretext training.}
    \label{fig:pretext_conf} \vspace{-10pt}
\end{figure}

Fig.~\ref{fig:pretext_conf} shows that the zero-shot probe with the model trained only on H3.6M, consistently over-predicts class 0 and fails on more severe classes. Fine-tuning on healthy gait helps separate classes 1 and 2 but still misses class 3. Only after fine-tuning on CARE-PD does the model begin to accurately identify all four classes, including class 3, reflecting better sensitivity to clinical severity.

\vspace{-10pt}\subsection{Subgroup Sensitivity}
    We evaluated whether predicted UPDRS-gait scores from finetuned MotionAGFormer reflected clinically meaningful subgroup differences. Medication analysis used BMClab and E-LC, FoG comparison used BMClab, KUL-DT-T and E-LC, and PD vs. healthy control analysis used DNE.

    \noindent\textbf{Medication.} Predicted UPDRS-gait scores were significantly lower when participants were on medication (median = $1.0$, IQR = $1.0$) compared to off-medication (median = $2.0$, IQR = $2.0$). This difference was statistically significant (Mann–Whitney U test, $p \leq 10^{-5}$), with a medium effect size (Cliff’s $\Delta = 0.42$) indicating that a random individual from group off-med has a 71\% chance of having a higher predicted UPDRS-gait score than a random individual from group on-med.
    These findings demonstrate the model’s sensitivity to medication status in assessing parkinsonian gait.

    \noindent\textbf{Freezing Status.} Predicted UPDRS-gait scores were significantly higher among freezers (median = $2.0$, IQR = $2.0$) compared to non-freezers (median = $1.0$, IQR = $2.0$). This difference was statistically significant (Mann–Whitney U test, $p \leq 10^{-5}$), with a moderate effect size (Cliff’s $\Delta = 0.25$), meaning a 62\% likelihood that a freezer has a higher predicted UPDRS-gait score than a non-freezer. This indicates the model is sensitive to capture increased gait impairment associated with FoG. 
    
    \noindent\textbf{Participant Type.} Predicted UPDRS-gait scores were significantly higher among participants with PD (median = $2.0$, IQR = $1.0$) compared to healthy controls (median = $1.0$, IQR = $1.0$). This difference was statistically significant (Mann–Whitney U test, $p \leq 10^{-5}$), with a large effect size (Cliff’s $\Delta = 0.50$), indicating that a randomly selected individual with PD has a 75\% chance of receiving a higher predicted UPDRS-gait score than a healthy control. These results support the model’s ability to distinguish between pathological and non-pathological gait patterns.
    
\begin{figure}[t]
        \centering
        \includegraphics[width=0.7\linewidth]{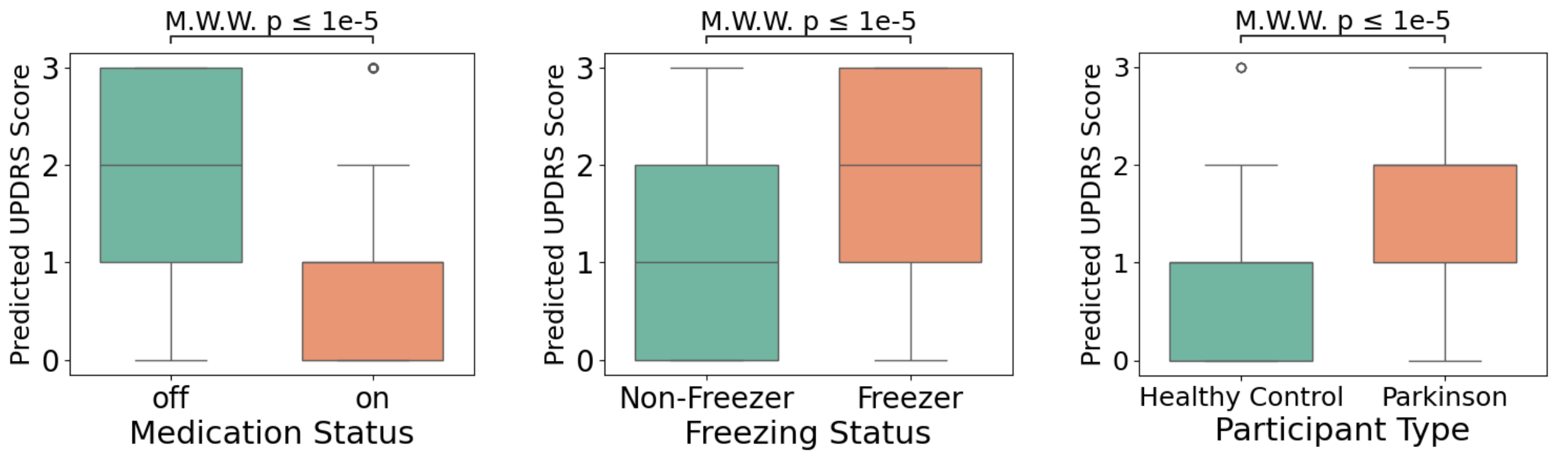}
        \caption{\small Predicted UPDRS-gait scores across three group comparisons: medication status (left), FoG status (middle), and diagnosis (right). M.W.W. refers to the Mann–Whitney–Wilcoxon test.}
        \label{fig:sensitivity} \vspace{-10pt}
    \end{figure}

    Figure~\ref{fig:sensitivity} summarizes these results, showing that the model consistently captures clinically meaningful group differences in predicted UPDRS-gait scores across all three sensitivity analyses.
    
\vspace{-10pt}\section{Conclusions}\vspace{-5pt}

We introduced \textsc{Care-PD}, a large multi-cohort 3D gait dataset for PD, enabling robust machine learning research through benchmark tasks in clinical score estimation and motion pretext learning. Results from seven backbone models and a handcrafted features baseline reveal four key lessons: 
First, pretrained encoders paired with classifier probe can capture some clinically relevant signals, but their accuracy collapses under distribution shifts—cross-dataset generalization is markedly harder, underscoring the need for multi-site evaluation. 
Second, dataset quality and size are as important as model architecture. Models trained or fine-tuned on larger cohorts generalize best, while those trained on smaller datasets underperform. Encoder choice also affects robustness to site variation.
Third, classical gait features are competitive within domains but less generalizable than learned representations.
Fourth, using \textsc{Care-PD} for pretext tasks improves both 3D estimation and downstream clinical prediction, confirming its value for both supervised and self-supervised clinical learning.
These findings motivate future work on clinical model development, domain-aware training, and test-time adaptation to capture pathological subtleties without overfitting site-specific biases. 
\textsc{Care-PD}’s scale and heterogeneity also give generative methods such as~\cite{adeli2025gaitgen} a strong foundation for generating more diverse, clinically grounded motion samples, helping address the scarcity of severe cases.
By providing data and evaluation protocols, \textsc{Care-PD} aims to accelerate the translation of motion AI into objective, scalable support for PD care.

\begin{ack}
This work was supported by the Data Sciences Institute at the University of Toronto, the Walter and Maria Schroeder Institute for Brain Innovation and Recovery and the AGE-WELL Network of Centres of Excellence (AGE-WELL NCE).
The authors sincerely thank these institutions for their support.

\end{ack}


\medskip

{
    \small
    \bibliographystyle{unsrt}
    \bibliography{main}
}

\newpage
\section*{Appendix}

\appendix
\renewcommand{\thefigure}{\thesection.\arabic{figure}}
\renewcommand{\thetable}{\thesection.\arabic{table}}
\setcounter{figure}{0}
\setcounter{table}{0}

\section{More Dataset Details}
\label{appendix:datasetdetail}
\begin{figure}[ht]
    \centering
    \includegraphics[width=1\linewidth]{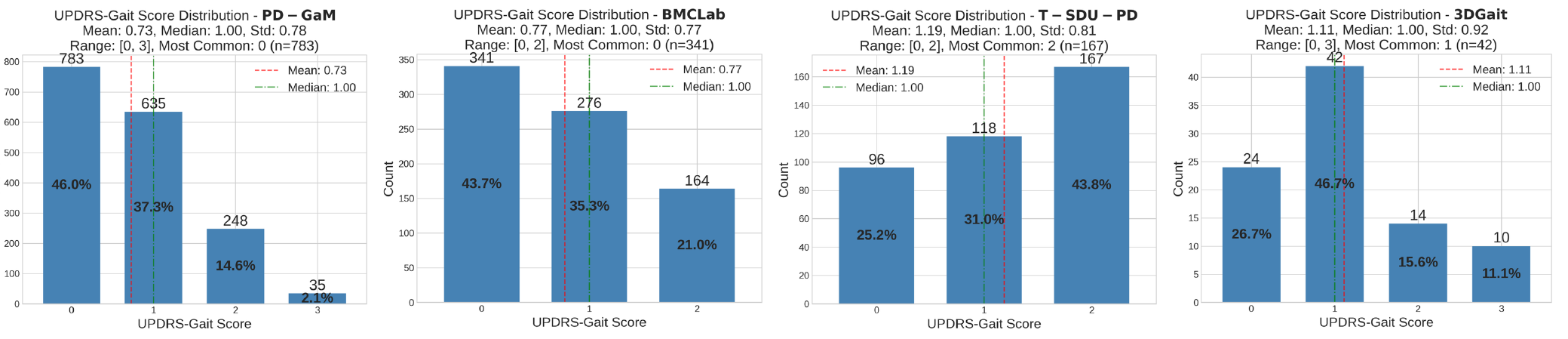}
    \caption{\small Class distributions for the four datasets with UPDRS-gait labels.}
    \label{fig:data_dist_updrs}
\end{figure}

\begin{figure}[ht]
    \centering
    \includegraphics[width=0.8\linewidth]{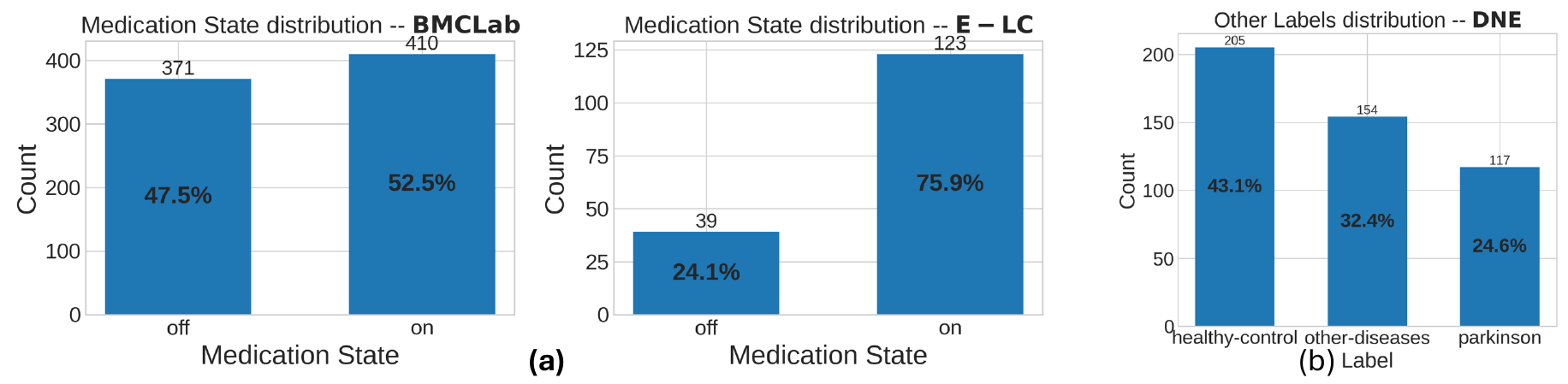}
    \caption{\small (a) Medication state breakdown for BMCLab and E-LC datasets. (b) Diagnostic categories in DNE dataset.}
    \label{fig:data_dist_med}
\end{figure}

\begin{figure}[ht]
    \centering
    \includegraphics[width=0.8\linewidth]{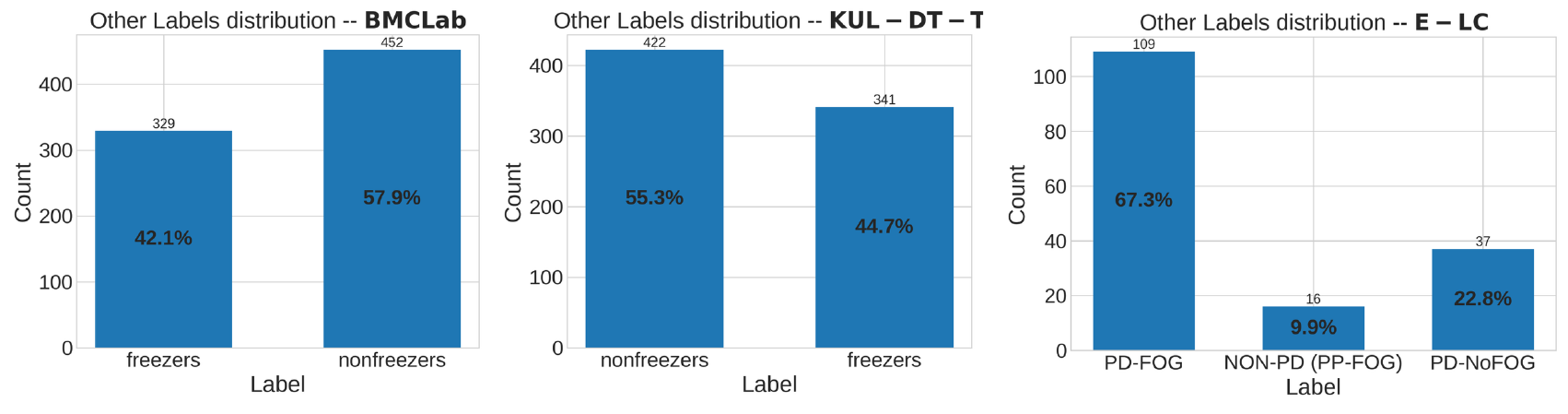}
    \caption{\small Label distributions for freezing status for BMCLab, KUL-DT-T and E-LC datasets.}
    \label{fig:data_dist_fog}
\end{figure}

Figure~\ref{fig:data_dist_updrs} presents the distribution of UPDRS-gait scores in the four labeled datasets. Score 0 (normal) is most common across cohorts, while score 3 (severe) is rare—especially in PD-GaM and 3DGait, highlighting class imbalance challenges. Figure~\ref{fig:data_dist_med} visualizes the distribution of (a)~medication states and (b)~diagnostic labels. BMCLab offers a balanced ON/OFF medication split, while E-LC is skewed toward ON-medication. DNE includes healthy, Parkinsonian, and other disease groups for broader contrastive training. Figure~\ref{fig:data_dist_fog} shows label distributions for FoG-related cohorts. BMCLab and KUL-DT-T distinguish freezers vs. non-freezers, while E-LC includes subtypes such as PD with FoG, PD without FoG, and non-PD with FoG symptoms.

\subsection{Slope Correction}
\label{appendix:slope}
\begin{figure}[t]
    \centering
    \includegraphics[width=\linewidth]{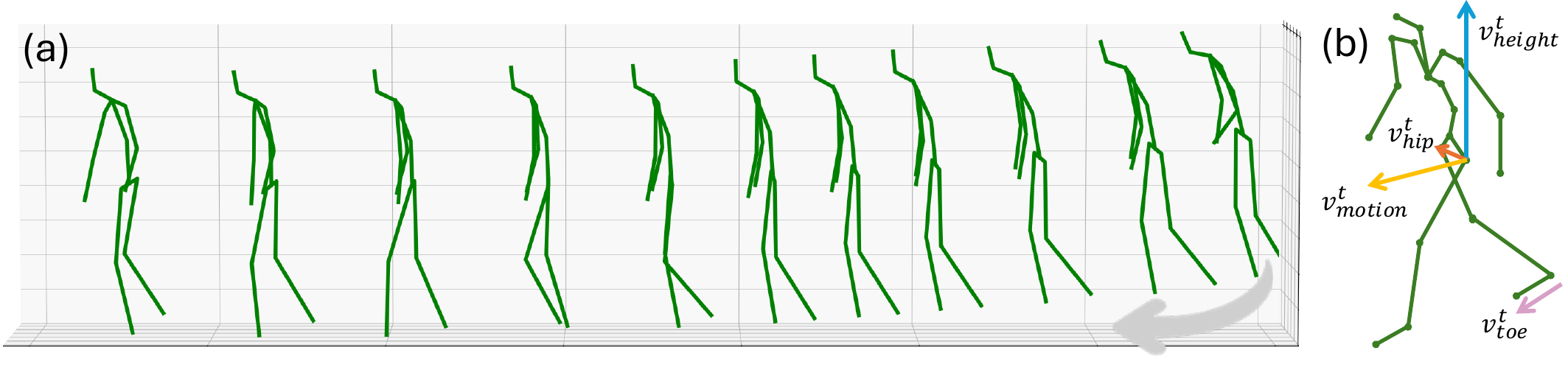}
    \caption{\small a) Illustration of the slope artifcat b) An example of the vertical height vector (blue), the direction of movement vector (yellow), and the hip vector (orange).}
    \label{fig:slope}
\end{figure}
In the \textsc{Care-PD} datasets recorded from ceiling-mounted cameras (T-SDU, T-LTC, and T-SDU-PD), we observed that sometimes subjects appeared to walk along a sloped or curved plane, rather than a flat floor. This artifact likely stems from the unusual top-down perspective—different from the front-facing or side views seen in WHAM’s training data~\cite{wham:cvpr:2024}. While motion encoder-based models may be robust to such distortions, feature-based gait classifiers rely on precise kinematic measurements and thus require carefully corrected input data.
To correct this slope artifact, we perform a frame-wise rigid alignment of the reconstructed SMPL skeleton using the Kabsch algorithm~\cite{lawrence2019purely}. The goal is to rotate each frame so that anatomical directions align with canonical coordinate axes (up, forward), while preserving natural gait structure.
Let the SMPL skeleton at time $t$ be a set of 3D joint positions:
\(
\mathbf{J}^t \in \mathbb{R}^{22 \times 3}.
\)
We define three key anatomical vectors per frame:

\paragraph{(1) Height Vector (posture):} defined as the offset between the sacrum and the average of the ankle and knee joint positions. 
\[
\mathbf{v}_{\text{height}}^t = \mathbf{j}^t_{\text{sacrum}} - \frac{1}{4} \left( \mathbf{j}^t_{\text{left ankle}} + \mathbf{j}^t_{\text{right ankle}} + \mathbf{j}^t_{\text{left knee}} + \mathbf{j}^t_{\text{right knee}} \right)
\]
This approximates the vertical posture and should align with the global $y$-axis: $\hat{\mathbf{y}} = [0, 1, 0]^T$. Misalignment suggests the subject appears tilted in 3D space.

\paragraph{(2) Motion Vector (walking direction):} To estimate walking direction, we compute the offset between the sacrum at frame $t$ and frame $t+15$, representing $\sim0.5$ seconds ahead. This motion vector is then projected onto the ground plane (xz-plane) and used as the walking axis. 
\[
\mathbf{v}_{\text{motion}}^t = \text{Proj}_{xz} \left( \mathbf{j}^{t+15}_{\text{sacrum}} - \mathbf{j}^t_{\text{sacrum}} \right)
\]
where $\text{Proj}_{xz}(\cdot)$ zeroes out the $y$-component. In frames where the sacrum displacement is less than 4mm—indicating near-stationary posture—we fall back on a proxy direction: the cross product of the hip vector (left hip to right hip) and the vertical vector. This gives a third perpendicular vector — ideally pointing forward along the walking direction. 

\[
\mathbf{v}_{\text{motion}}^t = \mathbf{v}_{\text{hip}}^t \times \mathbf{v}_{\text{height}}^t, \quad\quad \text{If}\; \| \mathbf{v}_{\text{motion}}^t \| < 4\text{mm}
\]
This proxy is adjusted to ensure consistency with foot orientation (by checking the sign of its dot product with toe direction and flipping the fallback direction when). We ensure alignment by flipping the fallback direction when
\[
\text{sign} \left( (\mathbf{v}_{\text{motion}}^t)^\top \cdot \mathbf{v}_{\text{toe}}^t \right) < 0, \quad\quad
\mathbf{v}_{\text{toe}}^t = \mathbf{j}^t_{\text{toe}} - \mathbf{j}^t_{\text{heel}}
\]

We normalize and smooth $\mathbf{v}_{\text{motion}}^t$ over time using a Savitzky–Golay filter~\cite{savitzky1964smoothing} (window=90, order=4) to ensure temporal coherence.

\paragraph{(3) Hip Vector (rotation anchor):} We assigned different importance to different pairs of vectors that should be aligned in the Kabsch algorithm. We set the weight for the alignment of the hip vector to infinity while the other two alignments were given a weight of 1. Thereby, we forced the hip vector ($\mathbf{v}_{\text{hip}}^t = \mathbf{j}^t_{\text{right hip}} - \mathbf{j}^t_{\text{left hip}}$) to stay aligned perfectly with itself while the other two vectors were allowed to deviate slightly from their targets. This prevents the correction from introducing unnatural body twisting to the subject’s gait.
Let
\[
\mathcal{S} = \{ (\mathbf{v}_i, \hat{\mathbf{v}}_i, w_i) \}
\]
where $\mathbf{v}_i \in \{\mathbf{v}_{\text{height}}^t, \mathbf{v}_{\text{motion}}^t, \mathbf{v}_{\text{hip}}^t\}$, target $\hat{\mathbf{v}}_i \in \{\hat{\mathbf{y}}, \hat{\mathbf{z}}, \mathbf{v}_{\text{hip}}^t\}$, and weights $w_i \in \{1, 1, \infty\}$.
We solve the weighted orthogonal Procrustes problem:
\[
\mathbf{R}^t = \arg\min_{\mathbf{R} \in SO(3)} \sum_i w_i \left\| \mathbf{R} \mathbf{v}_i - \hat{\mathbf{v}}_i \right\|^2
\]

The solution $\mathbf{R}^t$ is the optimal rotation aligning anatomical directions. We then apply this rotation to the entire skeleton around the root joint (sacrum) and translate the rotated skeleton vertically so that the lowest foot joint rests at 
$y=0$, ensuring ground contact consistency. This method corrects the slope artifacts while preserving the gait dynamics and anatomical validity of each sequence. An illustration of the process and vector definitions is shown in Fig.~\ref{fig:slope}.

\begin{figure}
    \centering
    \includegraphics[width=0.7\linewidth]{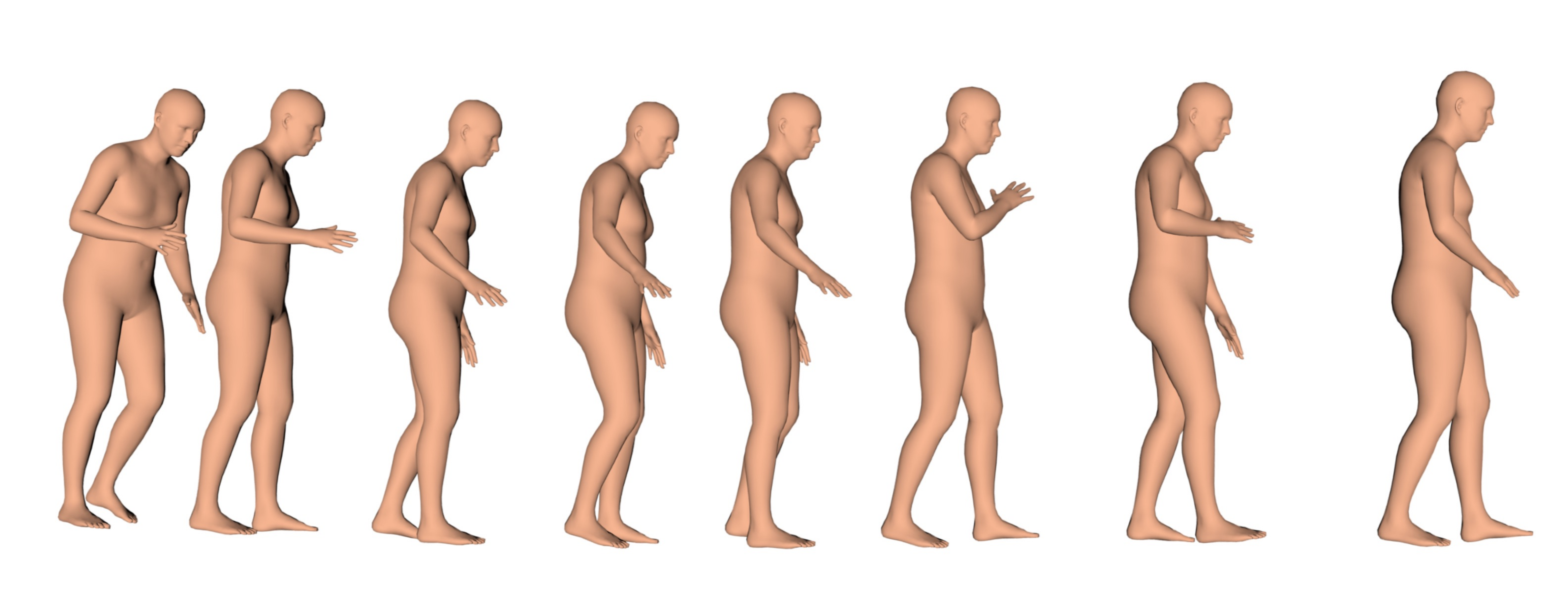}
    \caption{Example of the 6890 vertices SMPL mesh at different frames of the gait sequence.}
    \label{fig:smpl}
\end{figure}

\subsection{Clinical Validity of WHAM}\label{app-validity}
\begin{figure}
    \centering
    \includegraphics[width=1\linewidth]{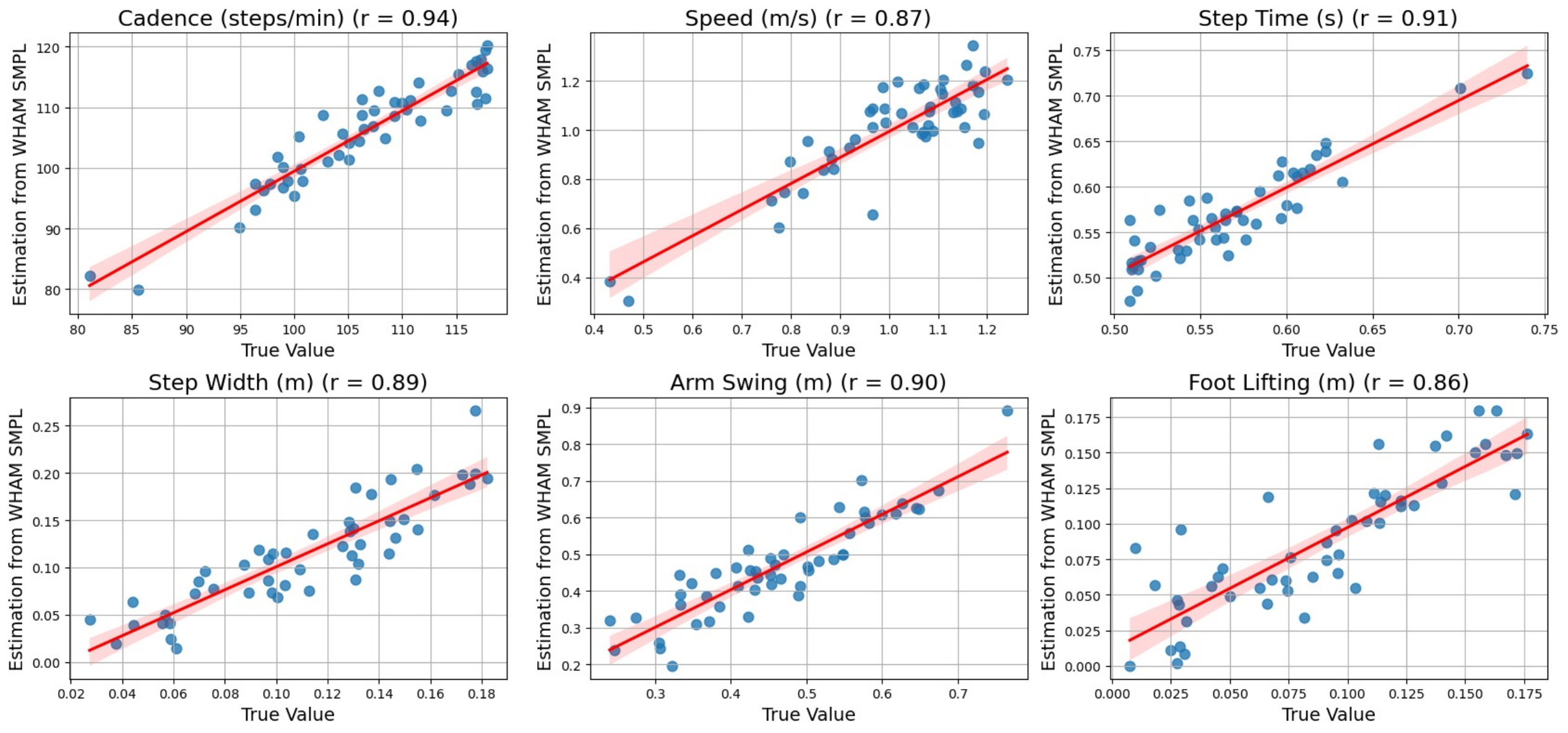}
    \caption{\small Correlation between WHAM estimations and IMU ground-truth gait features. Pearson correlations ranged from $r=0.86$ to $r=0.94$.}
    \label{fig:corr}
\end{figure}

We validated the WHAM SMPL estimations against a publicly available video + IMU benchmark, the Toronto Older Adults Gait Archive (TOAGA)~\cite{mehdizadeh2022toronto}. For each walk (from 14 participants) we extracted cadence, walking speed, step time, step width, arm-swing amplitude and foot-lifting height from the WHAM meshes and compared them to the features from their synchronised Xsens MVN Analyze 3D IMU recordings, which incorporate a reliable biomechanical model. Pearson correlations ranged from $r=0.86$ to $r=0.94$ (Fig.~\ref{fig:corr}), closely matching the high correlation originally observed between 2D pose-tracking and IMU measures in the TOAGA paper, supporting the biomechanical and clinical reliability of estimations. 
Furthermore, to quantify geometric accuracy we computed root-relative MPJPE between WHAM key-points and synchronised Xsens ground truth in TOAGA. The mean error was 39 mm, comfortably within the 35–65 mm range reported for multi-view pose estimation systems~\cite{joo2015panoptic, avogaro2023markerless, carrasco2022evaluation} and the TULIP dataset for PD gait task~\cite{kim2024tulip}. This level of agreement, together with high spatiotemporal feature correlations (Fig.~\ref{fig:corr}), supports the use of WHAM meshes as a reliable surrogate for markerless gait analysis in our study.

\subsection{Baseline Models and Baseline-specific Data Preprocessing}
\label{app:baseline_preproc}
To ensure that every backbone receives input in the format and frame-rate it was trained on, we applied a unified preprocessing pipeline along with baseline-specific preprocessing steps. All motion sequences were converted to 30~FPS to match the expected input frequency of the pretrained encoders. All preprocessing steps and motion representation generation procedures are available in our public code repository.

\subsubsection{Motion Formats}
\label{app:motion_formats}
To facilitate rigorous evaluation of motion encoder performance in clinical gait settings, we selected state-of-the-art models that operate on two broad classes of motion representations: skeleton-based (joint locations) and mesh-based (SMPL parameters). The SMPL-based models use either raw pose parameters or a redundant representation optimized for motion generation tasks. All formats are derived from SMPL as described below.
\paragraph{SMPL}
The SMPL model~\cite{loper2023smpl} represents human body shape $\beta$ and pose $\theta$ using 24 joints and a 10-dimensional shape vector. The pose is expressed as a set of joint rotations (e.g., axis-angle), and can be rendered as a mesh with 6890 vertices, an example of which can be seen in Fig.~\ref{fig:smpl}. For each time step \( t \), the SMPL input sequence \( \mathbf{M}^{1:T} \) has shape \( \mathbb{R}^{T \times 24 \times D} \), where \( D \) is the dimension of the rotation representation. SMPL serves as the base representation for generating other formats.
\paragraph{Human3.6M Joints}
Many encoders in our study were originally trained on the Human3.6M dataset~\cite{ionescu2013human3}, which uses a 17-joint skeleton. We project SMPL mesh vertices to this joint format using a linear regressor matrix \( \mathbf{R} \in \mathbb{R}^{17 \times 6890} \), as done in MotionBERT~\cite{zhu2023motionbert}. For each frame, the 3D Human3.6M joint coordinates are computed by multiplying the mesh with this regressor. The resulting motion sequence has shape \( \mathbb{R}^{T \times 17 \times 3} \).
\paragraph{HumanML3D}
The HumanML3D representation~\cite{guo2022generating}, originally introduced for text-to-motion generation, encodes each frame \( \mathbf{m}^t \in \mathbb{R}^{263} \) as a tuple of interpretable features \( \mathbf{m}^t=\{\dot{r}_a, \dot{r}_x, \dot{r}_z, r_{y}, \mathbf{j}_p, \mathbf{j}_v, \mathbf{j}_r, \mathbf{c}_f\} \), where
\( \dot{r}_a \in \mathbb{R} \), is the root joint’s angular velocity along the y-axis; \( \dot{r}_x, \dot{r}_z \in \mathbb{R} \), are the root’s linear velocities in the xz-plane; and \( r_y \in \mathbb{R} \), is the vertical height of the root joint. Joint-level features include \( \mathbf{j}_p \in \mathbb{R}^{3(N_j - 1)} \), the 3D positions of all joints except the root (21 joints); \( \mathbf{j}_v \in \mathbb{R}^{3N_j} \), the linear joint velocities; and \( \mathbf{j}_r \in \mathbb{R}^{6(N_j - 1)} \), the 6D joint rotations relative to parent joints in the skeletal hierarchy. Finally, \( \mathbf{c}_f \in \mathbb{R}^4 \) encodes four binary foot contact indicators derived from heel and toe velocities. This representation was computed from SMPL joints using the procedure introduced in ~\cite{guo2022generating}, yielding input tensors of shape \( \mathbb{R}^{T \times 263} \).

\subsubsection{Models}
\label{appendix:models}
Our benchmark intentionally spans \emph{diverse pre-training objectives, input formats, and architectural choices} so that conclusions about clinical transfer do not depend on a single modelling paradigm. To assess the clinical utility of pretrained motion representations, we evaluate seven state-of-the-art encoders spanning a range of architectures, training objectives, and input formats. These models were selected for their strong performance on benchmark motion tasks such as 2D to 3D lifting, motion reconstruction, prediction, and generation. They cover both skeleton-based and mesh-based representations and include both discriminative and generative paradigms. All models are used as fixed backbones; we extract their latent representations from last layer before final head (pooled over temporal dimension) and train lightweight classifiers on top for UPDRS-gait severity prediction.

\vspace{-7pt}\paragraph{POTR~\cite{martinez2021pose}}
is a transformer-based model originally developed for non-autoregressive human pose forecasting. Although designed for forecasting, its encoder learns strong spatiotemporal representations of input motion sequences shown to be useful for clinical gait assessment task~\cite{endo2022gaitforemer}. We use the encoder’s temporally pooled token embeddings as input features for our downstream clinical classifier. 
Input: 3D Human3.6M joints.

\vspace{-7pt}\paragraph{MixSTE~\cite{zhang2022mixste}}
is a 2D to 3D joint lifting model that factorizes spatial and temporal dependencies using stacked blocks of transformer encoders. Each block in its stacked architecture consists of a spatial transformer that captures joint-to-joint relationships within a single frame, followed by a temporal transformer that models how each joint evolves across time. Input: 2D projected (prespective) Human3.6M joints.

\vspace{-7pt}\paragraph{PoseFormerV2~\cite{zhao2023poseformerv2}} 
is a transformer-based model for 2D to 3D lifting that addresses two key challenges: computational efficiency and robustness to noisy 2D inputs. It applies a Discrete Cosine Transform (DCT) to each joint trajectory to obtain a compact, frequency-domain representation of global motion. Only a subset of low-frequency DCT coefficients are retained, effectively reducing noise from 2D pose estimators and shortening the sequence length. A spatial transformer encodes relations among joints using a fixed number of central frames, while the frequency features are linearly projected and concatenated with the spatial output. This combined representation is processed by a temporal transformer to model motion dynamics, and finally decoded back to the time domain. This architecture allows the model to capture long-range dependencies with reduced computational cost. Input: 2D projected (prespective) Human3.6M joints.

\vspace{-7pt}\paragraph{MotionBERT~\cite{zhu2023motionbert}}
is a dual-stream spatiotemporal transformer designed for 2D to 3D pose lifting. It takes 2D joint sequences as input and learns representations that capture both spatial relations among joints and temporal dynamics across frames. The model consists of stacked transformer blocks, each with two parallel branches: one applies multi-head self-attention in a spatial-first order (joint-wise attention followed by temporal), and the other in a temporal-first order. This design allows MotionBERT to learn complementary patterns in human motion while retaining frame-wise features useful for action recognition. The outputs from both streams are merged using a learned weighted average. In our setting, the final representation is obtained by averaging the output tokens across time. Input: 2D projected (orthographic) Human3.6M joints.

\vspace{-7pt}\paragraph{MotionAGFormer~\cite{mehraban2024motionagformer}}
extends the dual-stream transformer design of MotionBERT by integrating Graph Convolutional Networks (GCNs) into one of the branches. One stream uses MHSA to capture long-range dependencies, while the other applies spatial and temporal GCNs to model local joint interactions. The spatial GCN encodes the human body structure, while the temporal GCN builds connections based on feature similarity across time. This hybrid attention–graph architecture enhances robustness to localized variations in movement. Final features are obtained by temporally averaging the outputs across frames. Input: 2D projected (orthographic) Human3.6M joints.

\vspace{-7pt}\paragraph{MotionCLIP~\cite{tevet2022motionclip}}
is a transformer-based motion autoencoder trained for text-to-motion generation. During training, its latent space is aligned with the CLIP embedding space, enabling it to bridge motion and language domains; yet its motion encoder is a strong semantic aggregator. Including it tests whether language-aligned features, which never saw clinical labels, can be transferred to severity scoring. The model encodes SMPL pose sequences using stacked transformer layers and reconstructs them from the latent representation. For our experiments, we use its motion encoder as a frozen backbone and extract frame-level representations by averaging token outputs. MotionCLIP requires SMPL input in 6D rotation format, which avoids discontinuities associated with axis-angle representations and improves learning stability~\cite{zhou2019continuity}. Input: SMPL (6D rotation).

\vspace{-7pt}\paragraph{MoMask~\cite{guo2024momask}}
is a Vector Quantized VAE (VQ-VAE) based framework for text-conditioned 3D motion generation. It comprises a Residual VQ-VAE encoder–decoder for motion reconstruction and representation learning plus two transformers: a masked transformer for predicting base-layer motion tokens, and a residual transformer for refining higher-layer tokens. Unlike standard VQ-VAEs, MoMask uses multiple codebooks to iteratively quantize the residuals, enabling finer motion detail. We use the pretrained RVQ-VAE encoder as a feature extractor and obtain motion representations by summing tokens across all residual layers and averaging over time. MoMask operates on the HumanML3D representation and requires normalized features; normalization statistics are computed per dataset or per LODO training split. Input: HumanML3D.

\subsubsection{2D Projection Pipeline}
\label{app:projection}
To evaluate motion encoders trained on 2D joint data, we converted every 3D sequence into the appropriate 2D format via projection. To ensure a fair comparison between 2D and 3D models, given that projection discards depth information, we defined a multi-view setup for 2D encoders, using both back views, which minimize limb occlusion, and side views, which better preserve stride length. 
Using ground-truth 2D skeletons isolates an encoder’s \emph{representation capacity} from the performance of upstream key-point detectors and avoids confounds introduced by varying video quality across the eight sites.

The projection pipeline involves several steps: 
\quad
1)~Canonicalizing orientation of each regressed SMPL pose so that the initial walking direction faces $+z$. 
\quad
2)~\textit{Perspective projection}. We render ``perfect'' skeletons using two virtual pinhole camera models, viewing the walk from side and back\footnote{For additional implementation details of the projection setup, including camera configuration and rendering, refer to \texttt{data/preprocessing/smpl2h36m.py} in our codebase.}. \textit{Orthographic projection} for MotionBERT and MotionAGformer by removing the $z$ axis in camera coordinate. Views from the side and back were chosen to reflect common clinical perspectives.
\quad
3)~Pruning out-of-frame projections. Frames in which any joint projects outside the image plane are discarded. Also, sequences shorter than 30 frames after clipping are excluded.

To test whether complementary viewpoints help severity scoring, we build a ``Side \& Back'' variant: a Side (lateral) probe and a Back (posterior) probe are trained independently on their respective projections and their softmax outputs are averaged at inference time. All 2D encoder results reported in the manuscript use this multi-view fusion setup, as it consistently outperforms either Side or Back views alone.

\subsubsection{Input Normalizations}
\label{app:normalization}
For each model we followed its original preprocessing scheme.
For MixSTE and PoseFormerV2, input 2D joint coordinates were re-scaled to \([-1,1]\) in image space. For MotionAGFormer and MotionBERT cropping and rescaling normalization is used. Specifically, valid joint coordinates in the 2D image plane are tightly cropped to the bounding box of the motion, then linearly rescaled to the \([-1,1]\) range. The scaling is performed independently per clip using the larger of the height or width of the bounding box to preserve aspect ratio. POTR, which operates on 3D joint coordinates, centers each pose (i.e., per-frame joint set) on the pelvis and applies z-score normalization from the training set. MotionCLIP expects SMPL rotations in continuous 6D form; we therefore convert every axis-angle in the walk to 6D. For MoMask, we computed per-dataset mean/std (or, in LODO, mean/std on the pooled training sets) and divided the std of four root-velocity channels by a factor of 5, as recommended by the authors to emphasize global trajectory~\cite{guo2024momask}. For all the encoders, if a motion clip is shorter than the required input length, zero-padding is applied and a binary mask is used to track valid (non-padded) frames. For PoseFormerV2, which processes the central frames through a spatial transformer, we apply symmetric padding to preserve the alignment of meaningful motion content with the model’s receptive field.

\subsubsection{Generality of \textsc{Care-PD}.}
While our benchmarks focus on widely used motion formats and pretrained encoders, \textsc{Care-PD} is not restricted to these configurations. Its unified SMPL representation enables future work to explore other input types as well as specialized model architectures tailored to clinical gait analysis. We therefore view the present baselines as a starting point: future work can freely experiment with new motion formats and model classes that may prove even better suited to clinical gait analysis.

\subsection{Data Access and Preparation}

\begin{figure}[t]
    \centering
    \includegraphics[width=0.8\linewidth]{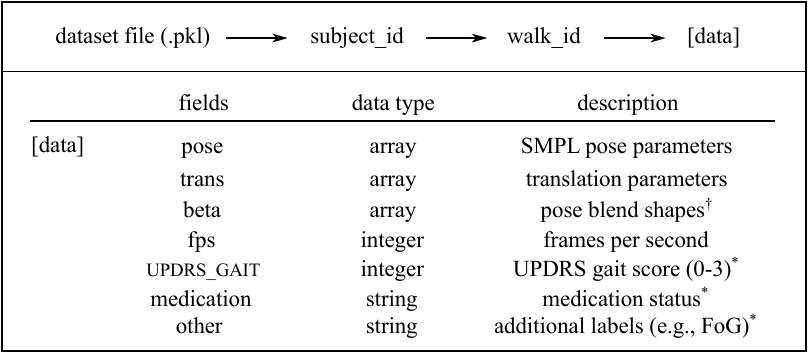}
    \caption{\small Each dataset within \textsc{Care-PD} is provided as a single .pkl data file, structured as illustrated. \textsuperscript{$\dagger$}Pose blend shapes are set to zero to preserve anonymity. \textsuperscript{*}Label information varies by dataset and is explicitly set as None if unavailable.}
    \label{fig:dataset_details}
\end{figure}

The \textsc{Care-PD} database is publicly accessible via the University of Toronto Dataverse. It is hosted by the University of Toronto Libraries, with data storage provided by the Ontario Library Research Cloud, a secure and geographically distributed cloud storage network developed in collaboration with partner universities across Ontario, Canada. The database is released under a CC-BY-NC license, allowing for open but non-commercial use with appropriate attribution. Detailed instructions for accessing the database can be found directly on the Dataverse project page~\href{https://doi.org/10.5683/SP3/TWIKMK}{\texttt{(Data)}} and the GitHub code base~\href{https://github.com/TaatiTeam/CARE-PD}{\texttt{(CARE-PD)}}. The structure of the \textsc{Care-PD} database’s metadata and SMPL data is visualized in Fig. \ref{fig:dataset_details}.
In addition to the SMPL data, \textsc{Care-PD} includes three derived assets to facilitate ease of use: Human3.6M, HumanML3D, and SMPL6D formats. For more information on these derived assets, we refer users to supplementary documentation in Sec. \ref{app:motion_formats} and our GitHub code base.

\section{Gait Feature Extraction Details}
\label{gaitfeat-app}
To build an interpretable baseline for UPDRS-gait classification, we extract a set of clinically meaningful gait features from 3D joint trajectories in Human3.6M format. These features, inspired by established clinical guidelines and prior work~\cite{mirelman2016arm, watson2021use, sabo2020assessment}, span spatiotemporal, stability, and posture-related dimensions relevant to parkinsonian gait.
\vspace{-7pt}\paragraph{Heel Strike Detection.}
Accurate detection of heel strike events is necessary for estimating step-level features. We compute the Euclidean distance between the left and right ankle joints over time identifying local maxima that are at least 8 frames apart and have a prominence of at least 0.02. These peaks approximate the alternating steps and define the heel strike timestamps.
\vspace{-7pt}\paragraph{Extracted Gait Features.}
Following~\cite{ng2020measuring}, we compute the following gait features, using the detected heel strikes:

\begin{itemize}[left=0.5em, topsep=1pt]
\item \textit{Cadence}: steps per minute, based on the total number of detected heel strikes.

\item \textit{Step Length / Width / Time}: 
computed between consecutive heel strikes. Step length is the distance measured along the walking ($z$) axis, step width along the mediolateral ($x$) axis at the time of each detected heel strike, and step time as the duration between strikes. Both the mean and standard deviation of these values are calculated.

\item \textit{Walking Speed}: total sacrum displacement between first and last heel strike, divided by total time.

\item \textit{Estimated Margin of Stability (eMoS)}: computed as the minimum distance between the extrapolated center of mass (XCoM) and base of support (feet) along the mediolateral direction. The hip vector approximates this axis. We calculated both the minimum (capturing the most unstable moment) and the standard deviation across steps.

\item \textit{Foot Lifting}: the vertical range of ankle movement.

\item \textit{Stoop Posture}: defined as the forward-lean distance is the vertical displacement between neck and sacrum, projected onto the direction of walk.

\item \textit{Arm Swing}: horizontal displacement of the hand joints along the forward axis, after translating the sacrum to the origin to remove global motion.

\end{itemize}

To ensure consistency, all sequences are pre-aligned to a canonical coordinate system (z-forward, y-up, x-lateral). This alignment is critical for ensuring geometric consistency when computing direction-sensitive features such as step length, step width, and stoop posture.
Previous studies~\cite{kim2018gait, zanardi2021gait, yoon2019effects} have demonstrated the relevance of these
gait features to the severity of PD symptoms. A low cadence and short step length are characteristic of slowness of movement, one of the hallmark symptoms of PD. While narrower step width and lower eMoS values reflect stability issues~\cite{mehdizadeh2022toronto}. PD may also manifest as patients taking shorter steps, resulting in elevated cadence~\cite{zanardi2021gait}. Moreover, a stooped posture is commonly seen in PD and is directly associated with postural instability~\cite{yoon2019effects}.

We use a Random Forest classifier to map the extracted gait features to UPDRS-gait score classes. The model is trained and evaluated using the same data splits, evaluation metrics, and hyperparameter tuning strategy as the encoder-based models (detailed in Appendix~\ref{app:hyperparams}), ensuring a consistent comparison across representation-learning and handcrafted approaches.

\section{Reproducibility}\label{app:reproducibility}
The experiments in this work can be reproduced using our Github repository, available at this link: \href{https://github.com/TaatiTeam/CARE-PD/}{https://github.com/TaatiTeam/CARE-PD/}.
Steps for how to reproduce evaluation experiments are available in our code \href{https://github.com/TaatiTeam/CARE-PD/}{README.md} and \href{https://github.com/TaatiTeam/CARE-PD/blob/master/docs/dataset.md}{dataset.md}.

\vspace{-7pt}\paragraph{Compute resources.}
All clinical score estimation task experiments were conducted on one NVIDIA A40 GPU hosted on a HPC cluster and pretext task experiments were conducted on a single RTX6000 GPU. 
In pretext experiments, training MotionAGFormer for 50 epochs took approximately 15 hours, while MoMask required around 2 hours for 30 epochs.
All code are implemented in PyTorch. More information on dependencies can be found on the Github page, installation guideline. 
Hyperparameter tuning was performed using Optuna~\cite{akiba2019optuna} with 50 trials per model-dataset pair. In all the experiments best set of hyperparameters were found in the first \(\sim\)30 trials.

\vspace{-7pt}\paragraph{Hyperparameter Tuning Details} \label{app:hyperparams}
During classifier training, all encoder backbones were kept frozen. We trained only the classifier head and tuned its hyperparameters using 6-fold stratified cross-validation on the BMClab dataset. BMClab was chosen due to its large size, clean motion capture quality, and pre-extracted walking segments. Hyperparameter search was conducted using the Optuna framework~\cite{akiba2019optuna} to explore a wide range of options, including learning rate {\small$\{1e^{-2}, 1e^{-3}, 1e^{-4}, 1e^{-5}\}$}, batch size {\small$\{64, 128, 256\}$}, number of training epochs {\small$\{10, 20, 30, 50, 70\}$}, weight decay {\small$\{0, 0.001, 0.01\}$}, and loss type (weighted cross-entropy or focal loss). For focal loss, the $\gamma \in$ {\small$\{1, 2\}$} parameter were included in the search and $\alpha$ was set to one. 

The best-performing hyperparameters discovered on BMClab  were reused across all datasets (see Fig.~1 in the paper). However, the optimal number of epochs was selected individually per dataset to account for differences in dataset size. Json file for the best set of hyperparameters used for each experiment is available in our GitHub page (\href{https://github.com/TaatiTeam/CARE-PD/tree/master}{\texttt{Link}}) in \texttt{configs/best\_configs\_augmented}
 folder. All splits used in cross-validation were subject-disjoint and stratified by label to prevent data leakage and ensure robust estimates. The exact fold splits used for each dataset and evaluation protocol are provided in the \texttt{folds} directory of our data repository.

For the LODO and MIDA experiments, the classifier head hyperparameters were again tuned on the combined training set (train set of the target dataset plus all the other datasets excluding the target's test set), using the same Optuna-based approach. The same hyperparameter tuning procedure was applied to the pretext task experiments.

The Random Forest classifier used in the engineered-feature baseline was also tuned using 6-fold subject-stratified cross-validation on the BMClab dataset. The optimal configuration was then applied uniformly across all experiments.

\section{More Experimental Results}\label{app:app-exp}

\subsection{Cross‑site robustness vs. in‑site accuracy}\label{app:robust}
\begin{figure}[t]
    \centering
    \includegraphics[width=0.9\linewidth]{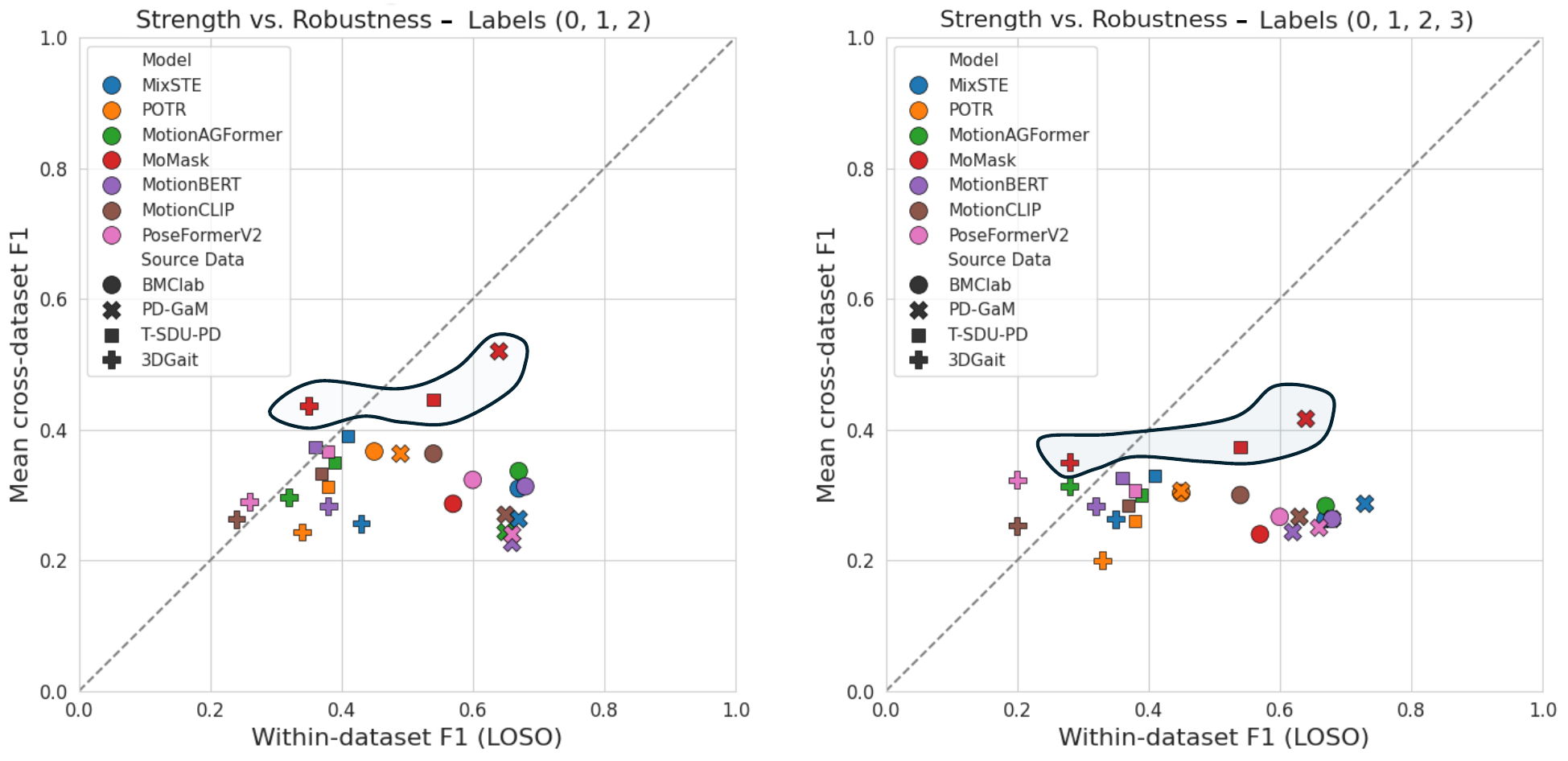}
    \caption{\small Accuracy vs. robustness analysis. Each marker represents an encoder plus linear prob trained on one dataset (marker shape) and evaluated on that dataset (x‑axis) and, on average, on the other three (y‑axis). Colours distinguish encoder backbones; The left plot reports macro‑$\text{F}1_{0-2}$, the right $\text{F}1_{0-3}$. The enclosed region highlights the most robust backbone and probes.}
    \label{fig:robustness}
\end{figure}

The two scatter plots in Fig.~\ref{fig:robustness} summarize every pair of ⟨encoder × source‑cohort⟩ probe by plotting its LOSO (within‑dataset) macro‑F\textsubscript{1} on the horizontal axis and the mean of its three off‑site scores on the vertical axis; the grey diagonal marks perfect transfer.
The left panel uses the 3‑class metric (labels 0‑2) whereas the right panel includes the rare severe class 3. 
The circled region highlights MoMask models that consistently combine strong within-dataset accuracy with robust cross-dataset generalization, with PD-GaM-trained variants showing the most prominent and reliable transferability, confirming that (i) breadth and heterogeneity of the source data are critical and (ii) this backbone make best use of that breadth.

Adding class 3 shifts every points trained on BMCLab and T-SDU-PD (the two dataset without label 3) downward, often by 5–10pp on the y‑axis, but the relative ordering is unchanged; models that were robust in the 3‑class setting remain the most robust once the challenging severe cases are re‑introduced. This pattern reinforces the earlier conclusion that scarcity of severe samples, is a major failure mode on cross‑site tests.

\subsection{Multi-dataset in-domain adaptation (MIDA) vs.\ baseline accuracy}
\begin{figure}[t]
    \centering
    \includegraphics[width=0.9\linewidth]{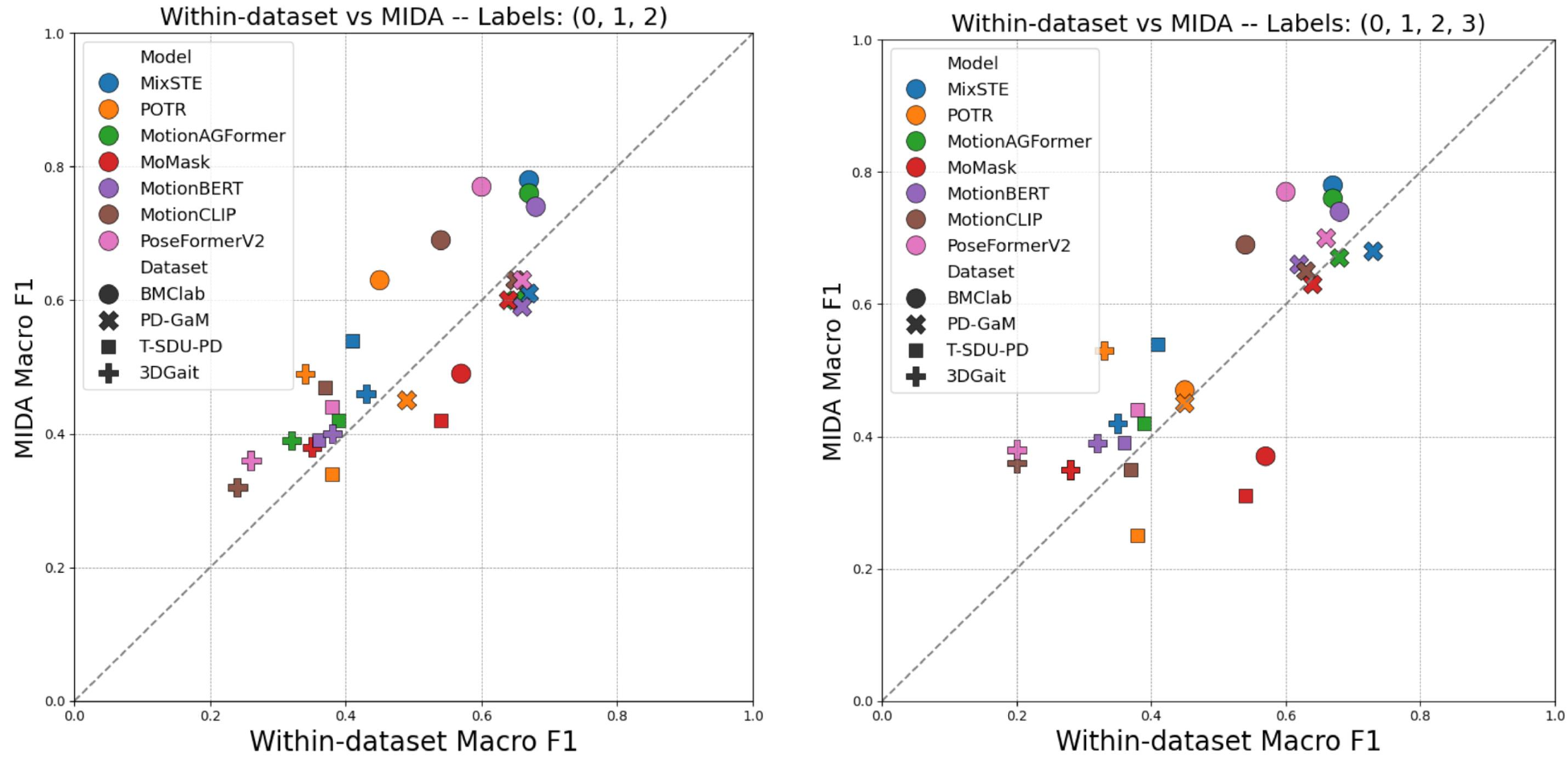}
    \caption{\small Within-dataset vs. MIDA evaluation (effect of in-domain adaptation and external data for training). Each point compares macro-F\textsubscript{1} scores of an encoder trained with (y-axis) and without (x-axis) access to additional out-of-domain datasets. LOSO uses training data only from the evaluation cohort; MIDA adds other datasets to training while still testing on the same held-out subjects. Colors indicate encoder backbone and shapes indicate target dataset.    
    The left panel uses macro-F\textsubscript{1} over labels 0–2, the right panel over 0–3. Markers above the diagonal indicate improvement; colours denote backbones, shapes the source cohort.}
    \label{fig:within_mida}
\end{figure}

Figure~\ref{fig:within_mida} contrasts standard LOSO evaluation (x-axis), where each model is trained solely on the target dataset, with MIDA (y-axis), where training includes both the target dataset and additional cohorts.
Most points rise above the diagonal, showing that supplementing a site’s own data with external cohorts usually helps, even though the test split is unchanged. 

\subsection{View-point results for 2D encoders}\label{app:view-ablation}
Table~\ref{tab:2d_all} reports the average macro-F\textsubscript{1} scores across datasets for four 2D models under the within- and cross-dataset, LODO, and MIDA evaluation protocols. 
We separately evaluated performance using posterior and lateral projections and their combination to assess the effect of viewpoint on model performance and robustness.
When only posterior or lateral projections are available, accuracy varies with the backbone. Fusing both views (“Combined’’) reliably boosts performance, suggesting that the two projections supply complementary depth cues.

\rowcolors{2}{gray!10}{white}
\renewcommand{\arraystretch}{1.1}
\setlength{\tabcolsep}{5pt}

\begin{table}[ht]
\centering\small
\caption{\small Average macro-F\textsubscript{1} (\%) of the 2D encoders across all datasets, grouped by evaluation protocol and viewpoint.  “Posterior’’ and “Lateral’’ use single-view projections, while “Combined’’ averages a posterior and a lateral probe at score level.  The upper half evaluates all four UPDRS classes, the lower half excludes the rare score 3.  Means (last column) are taken across the four backbones.}
\label{tab:2d_all}
\begin{tabular}{>{\columncolor{white}}llcccc|c}
\toprule
Protocol & View & MixSTE & MotionAGFormer & MotionBERT & PoseFormerV2 & Mean \\
\midrule
\multicolumn{7}{c}{\textit{Included Labels: \{0,1,2,3\}}}\\

       \cellcolor{white}  & Posterior & \textbf{35.19} & \textbf{35.69} & 25.81 & 28.87 & 31.39\\
 Within/Cross & Lateral   & 34.31 & 32.38 & \textbf{33.38} & \textbf{34.06} & \underline{\textbf{33.53}}\\
 \cellcolor{white}        & Combined  & \underline{34.94} & \underline{34.38} & \underline{\textbf{33.31}} & \underline{33.00} & \textbf{33.91}\\\midrule
\cellcolor{white}         & Posterior & 32.25 & \underline{34.00} & 25.75 & 30.00 & 30.50\\
\cellcolor{white}LODO         & Lateral   & \underline{33.01} & 33.75 & \underline{28.00} & \underline{30.50} & \underline{31.31}\\
\cellcolor{white}         & Combined  & \textbf{35.75} & \textbf{38.75} & \textbf{32.25} & \textbf{36.75} & \textbf{35.88}\\\midrule
\cellcolor{white}         & Posterior & \underline{55.75} & 39.75 & \underline{45.51} & \underline{54.25} & \underline{48.81}\\
\cellcolor{white}MIDA         & Lateral   & 39.75 & \underline{52.25} & 40.00 & 46.75 & 44.69\\
\cellcolor{white}         & Combined  & \textbf{60.51} & \textbf{55.67} & \textbf{54.51} & \textbf{57.25} & \textbf{56.99}\\
\midrule
\midrule
\multicolumn{7}{c}{\textit{Included Labels: \{0,1,2\}}}\\
\cellcolor{white} & Posterior & \underline{37.06} & \textbf{37.50} & 28.88 & 30.88 & 33.58\\
\cellcolor{white}Within/Cross & Lateral   & \textbf{37.38} & 34.19 & \underline{35.25} & \textbf{37.12} & \textbf{35.99}\\
\cellcolor{white} & Combined  & 36.51 & \underline{35.69} & \textbf{35.44} & \underline{34.75} & \underline{\textbf{35.60}}\\\midrule
\cellcolor{white}         & Posterior & 33.75 & \textbf{39.01} & \underline{29.01} & \underline{34.75} & \underline{34.13}\\
\cellcolor{white}LODO         & Lateral   & \textbf{37.25} & 33.25 & 28.75 & 33.00 & 33.06\\
\cellcolor{white}         & Combined  & \underline{35.75} & \underline{36.05} & \textbf{32.25} & \textbf{35.75} & \textbf{34.94}\\\midrule
\cellcolor{white}         & Posterior & \underline{55.75} & 43.75 & \underline{45.51} & \underline{52.25} & \underline{49.31}\\
\cellcolor{white}MIDA         & Lateral   & 45.25 & \underline{48.51} & 44.02 & 46.52 & 46.07\\
\cellcolor{white}         & Combined  & \textbf{59.75} & \textbf{54.25} & \textbf{53.08} & \textbf{55.07} & \textbf{55.54}\\
\bottomrule
\end{tabular}
\end{table}

\subsection{Variability Reporting}

We report mean macro-F\textsubscript{1} scores alongside their standard deviation 
to quantify variability and support statistical interpretation.
\emph{(i) LOSO:} inside each cohort we perform leave-one-subject-out; the 
$n$ held-out subjects yield $n$ scores. We report mean$\pm$SD across these 
$n$ folds.
\emph{(ii) Cross-dataset:} training on one cohort and testing on the other three gives $n=3$ off-site scores; the same formula provides mean ± SD.
\emph{(iii) MIDA:} we re-run LOSO after adding external data to the training split, so $n$ and the computation are identical to (i).
These statistics quantify, respectively, \emph{between-subject} and \emph{between-dataset} heterogeneity.

Table~\ref{tab:var_all} reports the resulting
mean$\pm$SD 
over all models.
Within-site LOSO yields the highest and most stable scores when the cohort itself is large and diverse (PD-GaM 62.0 $\pm$ 5.6~pp), but collapses on the small 3DGait set (27.1 $\pm$ 8.2~pp). Cross-site transfer is markedly harder: mean macro-F\textsubscript{1} drops by \textasciitilde25 pp on average, with wider confidence intervals, confirming that domain shift, is a major source of error. Adding auxiliary cohorts during training improves the accuracy in all the datasets. The persistent spread, however, shows that even with extra data the smaller or more idiosyncratic sites (T-SDU-PD, 3DGait) remain challenging, underscoring the importance of both scale and diversity in future clinical gait datasets.

\begin{table}[t]
  \centering\small
  \caption{\textbf{Between-subject and between-site variability.}
           Mean$\pm$SD macro-F\textsubscript{1} (\%), labels 0–3 over the seven encoders.
           }
  \label{tab:var_all}
  \rowcolors{2}{gray!10}{white}
  \setlength{\tabcolsep}{13pt}
  \begin{tabular}{lcccc}
    \toprule
           & \multicolumn{4}{c}{\textbf{Target dataset}}\\
    \cmidrule(lr){2-5}
    Protocol & BMClab & PD-GaM & T-SDU-PD & 3DGait\\
    \midrule
    \multicolumn{5}{c}{\emph{LOSO (within-site train and test)}}\\
    Mean F\textsubscript{1} & 55.9$\pm$13.6 & {62.0$\pm$5.6} & 41.7$\pm$5.2 & 27.1$\pm$8.2\\
    \midrule
    \multicolumn{5}{c}{\emph{Cross-dataset (train on source dataset test on target)}}\\
    Mean F\textsubscript{1} & 27.6$\pm$12.3 & 28.9$\pm$11.2 & {29.0$\pm$12.0} & 28.7$\pm$10.8\\
    \midrule
    \multicolumn{5}{c}{\emph{MIDA (LOSO: train on target train split + auxiliary datasets, test on target test split)}}\\
    Mean F\textsubscript{1} & 61.5$\pm$12.2 & {65.2$\pm$4.4} & 43.6$\pm$4.3 & 37.2$\pm$8.3\\
    \bottomrule
  \end{tabular}
\end{table}

\section{Ethics and Documentation}

\textsc{Care-PD} includes nine datasets, six of which are existing retrospective datasets that did not require new participant instructions. Ethical approval for use of these retrospective datasets was obtained from the Social Sciences, Humanities \& Education Research Ethics Board of the University of Toronto (REB \#47891). For the three newly collected datasets ethical approval was provided by the University Health Network Research Ethics Board (CAPCR ID 24-5835). Participants were informed clearly about the data acquisition process and provided informed
consent. All data were anonymized to protect participant identity and personal health information. The dataset is distributed under a CC-BY-NC research-only license to prevent misuse and ensure alignment with clinical and ethical standards. Detailed documentation supports transparency and reproducibility, and we expect \textsc{Care-PD} to drive clinically meaningful, generalizable machine learning research in PD assessment. Full ethical and procedural details can be found in the original publications for each dataset.

\section{Limitations and Broader Impact}\label{app:limitation}
While \textsc{Care-PD} represents a major step toward clinically grounded gait modeling, several limitations remain.

\textit{First}, despite its scale and diversity, the dataset remains imbalanced with respect to severe gait impairment (UPDRS-gait score 3), which is both clinically rare and difficult to capture due to mobility constraints. Future work may explore data augmentation or synthetic generation to address this gap. 
\textit{Second}, while the dataset covers diverse clinical environments and capture modalities, RGB recordings can introduce additional noise that may impact reconstruction quality. Although SMPL fitting and WHAM recovery have shown clinical utility, validated via TOAGA (~\ref{app-validity}), monocular errors in depth and distal-joint estimation may still affect downstream tasks.
Future releases could extend support from MoCap and RGB to wearable sensor modalities like IMUs to broaden compatibility and enable multimodal learning.
\textit{Third}, some datasets use the original UPDRS rubric, while others follow the revised MDS-UPDRS. While the two scales are largely compatible and map onto the same four severity levels, small wording and scoring adjustments, together with per-subject or per-session (rather than per-walk) annotations in several datasets, introduce additional label variability. Moreover, the UPDRS-III gait score was also found to have the highest inter-rater variability among all UPDRS-III scores, with an intraclass correlation coefficient of 0.746`\cite{de2019inter}.
\textit{Fourth}, all data are recorded in clinical corridors or labs; outdoor and in-home walking are absent.
\textit{Fifth}, our clinical evaluation focuses on gait severity classification; more fine-grained symptom estimation (e.g., stride irregularity, freezing episodes) is left for future work.
\textit{Finally}, while \textsc{Care-PD} provides a strong foundation for representation learning, clinical decision-making often requires temporal context across multiple visits or activities. Most datasets in \textsc{Care-PD} consist of single-task, short-segment gait walks; however, three of the cohorts (i.e., T-SDU, T-LTC, T-SDU-PD) include logitudinal recordings and could be explored in future work for temporal modeling.

Future releases will target richer labels (e.g. stride-level events, patient-reported outcomes), additional capture modalities, and semi-synthetic augmentation pipelines to balance class 3. As a future direction, we aim to release an identity-preserving, photorealistic video synthesis layer, turning the real videos into paired synthetic clips, so researchers can benchmark the entire video to clinical downstream pipeline end-to-end.
Despite these limitations, we believe \textsc{Care-PD} is a crucial step toward scalable, clinically meaningful motion AI. We encourage future work to build on its protocols and extend the dataset to even richer and more representative clinical populations.

\paragraph{Broader Impact}
Misuse of \textsc{Care-PD} is limited due to strict anonymization protocols detailed in Sec.~\textcolor{red}{3.2}. Nonetheless, improper training practices represent a potential misuse, particularly training models selectively on subsets biased towards certain demographics. For instance, there is an underrepresentation of women in the severe FoG PD datasets such as BMCLab, KUL-DT-T, and E-LC, each having more than 75\% male participants. Given this imbalance, caution should be exercised when extrapolating results. This underrepresentation of women in clinical FoG datasets is, however, a widely recognized phenomenon~\cite{tosserams2021sex}. More broadly, there is a risk that clinical decision-making could become overly reliant on automated predictions, which may fail to generalize to underrepresented subgroups if not carefully validated.

Despite these potential issues, the contributions of \textsc{Care-PD} toward advancing AI-driven gait analysis significantly outweigh the risks associated with its misuse, as long as clinical applications developed from \textsc{Care-PD} undergo thorough and independent validation. \textsc{Care-PD} has strong potential for positive societal impact: it enables scalable and objective assessments of Parkinsonian gait, encourages reproducibility through public release, and fosters standardization in a fragmented research area. To maximize impact and minimize harm, models developed using \textsc{Care-PD} should be rigorously validated in diverse clinical contexts.


\end{document}